\documentclass{article} % For LaTeX2e
\usepackage{iclr2023_conference,times}

\usepackage{amsmath,amsfonts,bm}

\def\eqref#1{equation~\ref{#1}}
\def\1{\bm{1}}

\DeclareMathAlphabet{\mathsfit}{\encodingdefault}{\sfdefault}{m}{sl}
\SetMathAlphabet{\mathsfit}{bold}{\encodingdefault}{\sfdefault}{bx}{n}

\usepackage[utf8]{inputenc} % allow utf-8 input
\usepackage[T1]{fontenc}    % use 8-bit T1 fonts
\usepackage{hyperref}       % hyperlinks
\usepackage{url}            % simple URL typesetting
\usepackage{booktabs}       % professional-quality tables
\usepackage{amsfonts}       % blackboard math symbols
\usepackage{nicefrac}       % compact symbols for 1/2, etc.
\usepackage{microtype}      % microtypography
\usepackage{xcolor}         % colors

\usepackage{graphicx}
\usepackage{amsmath,amssymb}
 \usepackage{mathrsfs}
\usepackage{bm}
\usepackage{subcaption}
\usepackage{pifont}

\usepackage{multirow}
\usepackage{makecell}
\usepackage{comment}
\usepackage{wrapfig}

\newcommand{\tcb}{\textcolor{blue}}
\newcommand{\tcgr}{\textcolor{gray}}

\newcommand{\etal}{\textit{et al}.~}

\newcommand{\ieno}{\textit{i}.\textit{e}.} %there is no space

\newcommand{\egno}{\textit{e}.\textit{g}.} %there is no space

\newcommand{\wrt}{\textit{w.r.t.}~}

\newcommand{\ourss}{HINP} %there is no "."
\newcommand{\oursf}{Hierarchical Implicit Neural Processes~} 
\newcommand{\oursfno}{Hierarchical Implicit Neural Processes} 
\title{Versatile Neural Processes for Learning Implicit Neural Representations}
\author{Zongyu Guo$^{1}$\thanks{Work done during an internship at Microsoft Research Asia.} 
, Cuiling Lan$^2$
, Zhizheng Zhang$^2$ 
, Yan Lu$^2$ 
, Zhibo Chen$^1$ \\
$^1$University of Science and Technology of China, $^2$Microsoft Research Asia \\
$^1$ \texttt{\{guozy@mail., chenzhibo@\}ustc.edu.cn} \\
$^2$ \texttt{\{culan, zhizzhang, yanlu\}@microsoft.com} \\
}

\iclrfinalcopy % Uncomment for camera-ready version, but NOT for submission.
\begin{document}

\maketitle

\begin{abstract}
Representing a signal as a continuous function parameterized by neural network (a.k.a. Implicit Neural Representations, INRs) has attracted increasing attention in recent years. Neural Processes (NPs), which model the distributions over functions conditioned on partial observations (context set), provide a practical solution for fast inference of continuous functions. However, existing NP architectures suffer from inferior modeling capability for complex signals. In this paper, we propose an efficient NP framework dubbed Versatile Neural Processes (VNP), which largely increases the capability of approximating functions. Specifically, we introduce a bottleneck encoder that produces fewer and informative context tokens, relieving the high computational cost while providing high modeling capability. At the decoder side, we hierarchically learn multiple global latent variables that jointly model the global structure and the uncertainty of a function, enabling our model to capture the distribution of complex signals. We demonstrate the effectiveness of the proposed VNP on a variety of tasks involving 1D, 2D and 3D signals. Particularly, our method shows promise in learning accurate INRs w.r.t. a 3D scene without further finetuning. Code is available \href{https://github.com/ZongyuGuo/Versatile-NP}{here}.

\end{abstract}

\section{Introduction}
%\tcr{For signal regression, a recent line of research has attracted increasing attention that models} signals (\egno, images, 3D scenes) as continuous functions.
%The continuous function maps \tcr{any given} input coordinates to the corresponding signal values. 
%A recent line of research \tcr{for signal regression} is to model signals (\egno, images, 3D scenes) as continuous functions that map the input coordinates into the corresponding signal values. %Such continuous mapping functions are implicitly defined by neural networks, 
A recent line of research on learning representations is to model a signal (\egno, image, 3D scene) as a continuous function that map the input coordinates into the corresponding signal values. %Such continuous mapping functions are implicitly defined by neural networks, 
By parameterizing a continuous function with neural networks, % (usually multi-layer perceptrons, MLPs), 
such implicitly defined representations, \ieno, implicit neural representations (INRs), offer many benefits over conventional discrete (\egno, grid-based) representations, such as the compactness and memory-efficiency \citep{sitzmann2020implicit,tancik2020fourier,mildenhall2020nerf,chen2021nerv}.
Characterizing/parameterizing a signal by a corresponding set of network parameters generally requires re-training the neural network, %\citep{sitzmann2020implicit,tancik2020fourier,mildenhall2020nerf,chen2021nerv}, 
which is computationally costly. 
%Some prior works \citep{sitzmann2020metasdf,tancik2021learned,lee2021meta} explore learning INRs with gradient-based meta-learning methods \citep{finn2017model}. However, these works still require a few steps of gradient descent \tcr{for adapting} to a new signal \tcr{in testing}. 
%However, these works require the computation of high-order derivatives during meta-training and still have a few steps of gradient descent when it comes to a new signal. 
In practice, at test time, it is desired to have models that support fast adaptation to partial observations of a new signal without finetuning.%, \ieno, approaching the continuous function of this signal . 
% In practice, at test time, it is desired to have models that support fast adaptation to \tcr{partial} observations of a new signal, \ieno, approaching the continuous function of this signal \tcr{without finetuning}. 
% \tcb{Some prior works \citep{sitzmann2020metasdf,tancik2021learned} explore learning INRs with gradient-based meta-learning methods. However, these works require the computation of high-order derivatives during meta-training and still have a few steps of gradient descent when it comes to a new signal. }

In fact, the Neural Processes (NPs) family \citep{jha2022neural} supports such merit. It meta-learns the implicit neural representations of a probabilistic function conditioned on partial signal observations. During test-time inference, it enables the prediction of the function values at target points within a single forward pass. 
%\tcb{Another path to meta-learn the implicit neural representations is to capture the uncertainty of function distributions and \textit{generate} appropriate function representations w.r.t. a specific signal. This kind of methods are members of the Neural Processes (NPs) family as they approximates the distribution in function space that forms a Stochastic Processes \citep{ross1996stochastic}.}
% \tcb{Another path to meta-learn the implicit neural representations is to capture the uncertainty of function distributions and \textit{generate} appropriate function representations w.r.t. a specific signal. This kind of methods are members of the Neural Processes (NPs) family as they approximates the distribution in function space that forms a Stochastic Processes \citep{ross1996stochastic}.}
Naturally, given partial observations of a signal, 
there exists uncertainty inside its continuous function since there are many possible ways to interpret these observations (\ieno, context set).
%there exists uncertainty inside its continuous function since there are many possible ways to interpolate these observations (\ieno, context set).
The NP methods \citep{garnelo2018conditional,garnelo2018neural} learn to map a context set of observed input-output pairs to a conditional distribution over functions (with uncertainty modeling). However, it has been observed that NPs are prone to underfit the data distribution. 
% The NP methods \citep{garnelo2018conditional,garnelo2018neural} learn to map a context set of observed input-output pairs to a distribution over functions (with uncertainty modelling) for fast adaptation. 
%\tcr{By conditioning on observed context set, these methods allow test time inference.}
% These methods allow (\uppercase\expandafter{\romannumeral1}) the incorporation of information from an observed context set and make predictions with a single forward pass, and (\uppercase\expandafter{\romannumeral2}) \tcr{probabilistic prediction of the signal through the uncertainty modeling.} 
%\tcb{better approximation and easy computation of function distributions compared with the Gaussian Processes \citep{rasmussen2003gaussian}.
% (by Lan) Latent NP \citep{garnelo2018neural} learns a latent Gaussian variable distribution to capture the global uncertainty in the overall structure of the function. 
Following the spirits of variational auto-encoders \citep{kingma2014auto}, the work of \citep{garnelo2018neural} introduces a global latent variable to better capture the uncertainty in the overall structure of the function, which still suffers from the inferior capability for modeling complex signals.
%From optimization perspective, NPs are similar to a Variational Autoencoder (VAE) \citep{}, where NPs estimate the conditional distribution over the labels of the target points given the set of labeled context points, VAE estimates the distribution over the dataset given a sample. 
%It has been observed that NPs are prone to underfit to the data distribution. 
Attentive Neural Processes (ANP) \citep{kim2019attentive} can further alleviate this issue, 
%s elf-attention is applied to the context points to compute representations of each point and cross-attentionthe target input attends to these context representations to predict the target output.
which leverages the permutation-invariant attention mechanism \citep{vaswani2017attention} to reweight the context points and the target predictions.
%self-attention is applied to the context points to compute representations of each point and cross-attention attends the target input to these context representations to predict the target output.} 
However, taking each context point as a token, ANP has troubles in processing complex signals that requires abundant context points as condition (\egno, image with high resolution), where the computational cost is very expensive. Moreover, for complex signals, modeling the global structure and uncertainty of the function with a single latent Gaussian variable may be suboptimal. It is worthwhile to explore an efficient framework to excavate the potential of NPs in modeling complex signals. 

\begin{figure*}[t]
 \centering
 \begin{subfigure}{0.98\linewidth}
\includegraphics[scale=0.55, clip, trim=3.8cm 11.5cm 5.1cm 1.2cm]{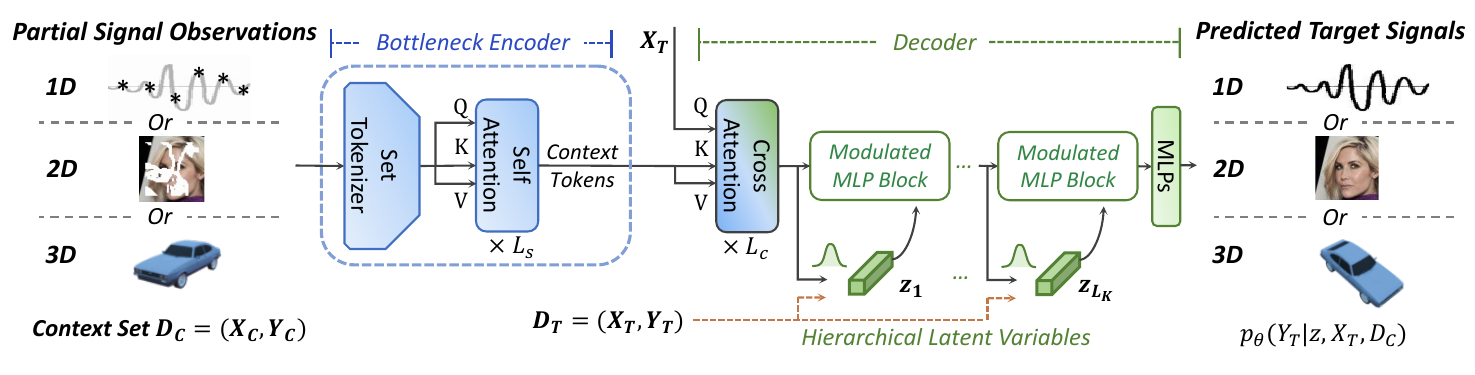}
 \end{subfigure}
  \caption{The proposed Versatile Neural Processes framework contains a bottleneck encoder and a hierarchical latent modulated decoder. The input context set is first encoded into fewer and informative context tokens by a set tokenizer followed by self-attention blocks, which provide powerful network capability with tolerable complexity. The decoder consists of cross-attention modules and multiple modulated MLP blocks, enhancing the model expressiveness for complex signals.}
%  \caption{The propose VNP framework largely increases the expressiveness and flexibility of function representations, better handling a variety of tasks on synthetic and real-world signals, including 1D function regression, 2D image completion and superresolution to arbitrary size, and few-shot novel view synthesize in 3D scenes. }
 % on CelebA64 \citep{liu2015deep}
\label{figure1}
\end{figure*}

In this paper, we propose Versatile Neural Processes (VNP), an efficient and flexible framework for 
%continuous function approximation. 
meta-learning of implicit neural representations.
%As shown in Figure~\ref{figure1}, VNP consists of a bottleneck encoder and a hierarchical latent modulated decoder. Specifically, the bottleneck encoder will \tcb{transfer} the set of context points into a reduced (informative) set of context tokens, alleviating the requirement of high computational cost. 
Figure~\ref{figure1} shows the framework of VNP. Specifically, VNP consists of a bottleneck encoder and a hierarchical latent modulated decoder. 
%The bottleneck encoder \tcr{encodes} the set of context points into fewer informative context tokens, refraining from high computational cost \tcr{especially} on complex signals. 
The bottleneck encoder powered by the set tokenizer and self-attention blocks encodes the set of context points into fewer and informative context tokens, refraining from high computational cost especially on complex signals while attaining higher modeling capability. 
% The bottleneck encoder will \tcb{transfer} the set of context points into fewer informative context tokens, \tcb{refraining from high computational cost on complex signals.
%\tcb{The set tokenizer and the subsequent attention layers ensure the capability and flexibility of this encoder in processing complex signals.}
At the decoder, we hierarchically learn multiple latent Gaussian variables for jointly modeling the global structure and uncertainty of the function distributions. Particularly, we sample from the latent variables and use them to modulate the parameters of the MLP modules.
Our VNP has high expressiveness to complex signals (\egno, 2D images and 3D scenes) and significantly outperforms existing NPs approaches on 1D synthetic data.

We summarize our main contributions as below:
%In short, our contributions are summarized as,
\begin{itemize}
%\item We propose an efficient NP framework, Versatile Neural Process (VNP), that is capable of approximating the functions of complex signals. 
% \item \tcb{We propose the Versatile Neural Processes (VNP) that is capable of learning accurate INRs of various signals within a single forward pass. }
\item We propose Versatile Neural Processes (VNP) that is capable of learning accurate INRs for approximating the function of a complex signal. 
%we decompose the implicit neural representations into local and global branches. We improve both branches and enable generating accurate implicit neural representations in a single forward pass.

\item We introduce a bottleneck encoder to produce compact yet representative context tokens, facilitating the processing of complex signals with tolerable computational complexity.
% \item \tcb{We introduce a bottleneck encoder to produce effective local features queried by target coordinates, improving both the capability and flexibility in processing complex signals.}

% by Lan
%\item We introduce a bottleneck encoder that facilitates tolerable computation complex signals by encoding the context set to a small set of context tokens. 

%\item We design HLMDec that jointly learn multiple global latent variables, which to enable the modeling of the distribution of complex function. 

\item We design a hierarchical latent modulated decoder that can better capture and describe the structure and uncertainty of functions through the joint modulation from the multiple global latent variables. 
%A hierarchical latent variable model is introduce to modulate the parameters of multiple global-shared MLPs which progressively enhance the signal value at any target location.
% \item A hierarchical latent variable model is introduce to modulate the parameters of multiple global-shared MLPs which progressively enhance the signal value at any target location.

\item We implement the VNP framework on 1D, 2D, and 3D signals respectively, demonstrating the state-of-the-art performance on a variety of tasks. Particularly, our method shows promise in learning accurate INRs of 3D scenes without further finetuning.
%\item We implement our proposed framework for diverse signal formats, demonstrating the state-of-the-art performance on a variety of tasks (See Figure \ref{figure1}). %, including 1D function regression. 2D image completion and superresolution, and few-shot novel view synthesize in 3D scenes.
\end{itemize}

\section{Related work}
\textbf{Implicit Neural Representations (INRs).} INRs aim at parameterizing a signal by a differentiable neural network, \ieno, learning a \textit{continuous} mapping function \wrt the signal~\citep{stanley2007compositional,sitzmann2020implicit,tancik2020fourier}. 
%INRs represent a \textit{continuous} mapping function \wrt the signal by a differentiable neural network.
%INRs aim at parameterizing a  signal by a differentiable neural network, which learn a continuous mapping function \wrt the signal.
%, \ieno, $ f_{\theta}: \mathcal{X} \rightarrow \mathcal{Y}$. 
In the seminal work CPPN \citep{stanley2007compositional}, a neural network is trained to learn the implicit function that fits a signal, \egno, an image.
Given any spatial position identified by a 2D coordinate, the model that acts as a function, outputs the color value of this position.  
Such continuous representations, as a powerful paradigm, have a wide range of applications such as image super-resolution \citep{chen2021learning}, modeling shapes \citep{chen2019learning,park2019deepsdf} and textures \citep{oechsle2019texture}, 3D scene reconstruction \citep{mildenhall2020nerf,martin2021nerf,niemeyer2021giraffe}, and even lossy compression \citep{dupont2021coin, dupont2022coin++, schwarz2022metalearning}. 
%Thanks to the continuous nature, INRs have many promising merits over conventional discrete representations, \egno, getting rid of the constraint of signal resolution and the memory-efficiency. 
Most of these methods require re-training the neural network to model/overfit a new signal, which is computationally costly \citep{sitzmann2020implicit,tancik2020fourier,mildenhall2020nerf,chen2021nerv, dupont2021coin}.

In practice, it is desired to have models
that support fast adaptation to a new signal, \ieno, approaching the continuous
function of this signal without abundant steps of optimization in inference. Some works \citep{chen2021learning,lee2021meta,sitzmann2020metasdf} adopt standard gradient-based meta-learning algorithms to learn the initial weight parameters of the network \citep{finn2017model}. However, they still require a few gradient decent steps to fit for the new signals. 
%ECCV
%There is a recent work \citep{chen2022transformers} that also applies transformer for learning INRs within a single forward pass. However, this work is designed on grids and cannot handle arbitrary context and target coordinates. We should note \citep{chen2022transformers} is not a probabilistic generative model and is thus conceptually different with the probabilistic solutions in the NPs family.
% \textbf{Fast Adaptation to New Signals.} In practice, it is desired to have models
% that support fast adaptation to partial observations of a new signal, i.e., approaching the continuous
% function of this signal within a short time. The work of \citep{chen2021learning} applies standard gradient-based meta-learning algorithms to learn the initial weight parameters of the network. Later, the work of \citep{lee2021meta} improves the meta-learned initial weights under a sparse constraint. These two works still require a few gradient decent step to fit for the new signals. There is a recent work \citep{chen2022transformers} that also applies transformer for learning INRs within a single forward pass. However, this work is designed on grids and cannot handle arbitrary context and target coordinates. We should note \citep{chen2022transformers} is not a probabilistic generative model and is thus conceptually different with the probabilistic solutions in the NPs family.

\textbf{Neural Processes (NPs).} Neural Processes actually can learn the continuous function conditioned on partial observations of a signal, enabling fast adaptation to a new signal (not requiring finetuning in inference). The series of NPs methods \citep{garnelo2018conditional,garnelo2018neural} provide probabilistic solutions in predicting continuous functions from partial observations. NPs approximate the distributions in the function space, which formulates Stochastic Processes \citep{ross1996stochastic}, by introducing stochasticity in function realization.
%They allow fast adaptation of the functions to the partial observations.
%approximating the uncertainty of a set of functions, 
%which are trained via meta-learning \citep{thrun2012learning}.
%Inference of the continuous function enables prediction of the unknown function values at arbitrary location, similar to meta learning \citep{}. 
A line of researches on neural processes has been introduced, targeting at improving the prediction accuracy \citep{kim2019attentive,lee2020bootstrapping,wang2020doubly,volpp2021bayesian}, preserving the stationarity of stochastic processes \citep{Gordon2020convolutional,foong2020meta}, and generalizing to observation noise \citep{kim2022neural}. 
The work of Neural Processes \citep{garnelo2018neural} learns a latent variable distribution to capture the global uncertainty in the overall structure of the function, which is optimized with variational inferece \citep{kingma2014auto}. 
%From optimization perspective, NPs are similar to a Variational Autoencoder (VAE) \citep{kingma2014auto}, where NPs estimate the conditional distribution over the labels of the target points given the set of labeled context points, VAE estimates the distribution over the dataset given a sample. 
Attentive Neural Processes (ANP) leverages the attention mechanism to enhance the representation of each context point and alleviate the underfitting problem \citep{kim2019attentive}. Transformer Neural Processes (TNP) \citep{nguyen2022transformer} similarly takes each context point as a token and leverages transformer architecture to approximate the function. However, for complex signals that requiring abundant context points as condition (\egno, image with high resolution), the computational complexities of ANP and TNP are very high which are quadratic with respect to the number of context points. 

It is desired to have a framework that can effectively approximate the functions of complex signals. In this work, we introduce a strong NP framework, Versatile Neural Processes (VNP), which leverages informative context tokens and explores the hierarchical global latent variables for modulation, leading to superior approximation of function distributions.

\section{Revisiting the Problem Formulation of NPs}

Neural Processes (NPs)  \citep{garnelo2018neural,garnelo2018conditional} are a class of methods that approximate the probabilistic distribution of continuous functions conditioned on partial observations.
% Neural Processes (NPs)  \citep{garnelo2018neural,garnelo2018conditional} are a class of methods that enable \tcr{test-time} rapid regression of a continuous function, which are trained to approximate the function distribution.
Suppose we have a labeled context set $D_C = (X_C, Y_C): = (\mathbf{x}_i, \mathbf{y}_i)_{i \in C}$ sampled from a continuous function with inputs $\mathbf{x}_i$ and outputs $\mathbf{y}_i$. NPs 
target at predicting arbitrary and finite target points $D_T = (X_T, Y_T) := (\mathbf{x}_i, \mathbf{y}_i)_{i \in T}$ by learning the input-output mapping function $f$. 
%, where $X$ and $Y$ are the function input and output.
% Suppose we have a labeled context set $D_C = (X_C, Y_C) = \{(x_i, y_i)\}_{i=1}^N$ ($N$ denotes the number of context points) sampled from a continuous function, NPs 
% target at predicting an arbitrary and finite number of target function values $D_T = (X_T, Y_T) := (x_i, y_i)_{i \in T}$, where $X$ and $Y$ are the function input and output. 
%Although a continuous function can actually have infinite sampling points, if the order in context set or target set is permutation-invariant like in NPs, it is able to approximate the distribution of continuous functions with finite sampling points \citep{oksendal2003stochastic}. 
Given some signal observations (a set of context points), many possible functions may match well to these observations and thus naturally there exists function uncertainties. The conditional distributions of targets points can be modeled as:
%The uncertainty of a continuous function, which implicitly determines a signal, can be measured with finite context and target points by taking a probabilistic stance:
\begin{equation}
   p_{\phi}(Y_T|X_T, D_c) = \prod\nolimits_{(\mathbf{x}, \mathbf{y}) \in D_T}{\mathcal{N}(\mathbf{y};\mu_\mathbf{y}(\mathbf{x},D_c), \sigma_\mathbf{y}^2(\mathbf{x},D_c))}.
\label{equation1}
\end{equation}
To generate coherent function predictions and better model the function distributions, Garnelo \etal \citep{garnelo2018neural} introduce the (Latent) Neural Process by encoding the global structure and uncertainty of the function into a latent Gaussian variable via Bayesian inference:
\begin{equation}
    \mathbf{z} \sim p_{\theta}(\mathbf{z}|X_T, D_C); \quad p_{\phi, \theta}(Y_T|X_T, \mathbf{z}) = \prod\nolimits_{(\mathbf{x}, \mathbf{y}) \in D_T}{\mathcal{N}(\mathbf{y};\mu_\mathbf{y}(\mathbf{z},X_T, D_C), \sigma_\mathbf{y}^2(\mathbf{z},X_T, D_C))}.
\label{equation2}
\end{equation}
Due to the intractable log-likelihood, some previous works adopt amortized variational inference \citep{kingma2014auto}, which is also used to optimize our proposed framework. We can derive the evidence lower bound (ELBO) on $\log p_{\theta}(Y_T|X_T, \mathbf{z})$, where the ELBO can be viewed as a combination of the reconstruction term (the first term) and the KL term (the second term):
\begin{equation}
    \mathbb{E}_{\mathbf{z} \sim q_{\phi}(\mathbf{z}|D_T)} [\log p_{\theta}(Y_T |  \mathbf{z}, X_T, D_C)] - D_{KL} [q_{\phi}(\mathbf{z}|D_T) || p_{\psi}(\mathbf{z}|X_T, D_C) ].
\label{equation3}
\end{equation}
Here, $\psi$ and $\theta$ refer to the parameters of conditional prior encoder and the decoder, similar to conditional VAEs \citep{sohn2015learning,ivanov2019variational}. $q_{\phi}(\mathbf{z}|D_T)$ denotes the posterior distribution of the latent variable given the ground truth target points, which is only used in training and not accessed in inference.
So far, we have revisited the problem formulation of NPs. Since NPs approximate the distribution over functions with some function observations, they can also be regarded as \textit{probabilistic, continuous, conditional} generative models, where the generated result is a continuous function. 

\section{Versatile Neural Processes}

We propose an efficient NP framework, Versatile Neural Processes (VNP), which provides  much improved capability in approximating the function of a various signal.
%for continuous function approximation conditioned on partial observation of a signal. 
Figure \ref{figure1} provides an overview of our proposed VNP.
%This is a type of regression problem. 
VNP is generic that can be used to generate the functions for 1D, 2D or 3D signals. For test-time inference, the inputs are a set of context points $D_C = (X_C, Y_C)$ and the coordinates of target points $X_T$. The outputs are the predicted values of target points $Y_T$. The continuous function is parameterized by the network weights.
%We have a few observations of the functions representing 1D, 2D or 3D signals.
%The proposed VNP consists of a bottleneck encoder that encodes the context points into fewer informative context tokens and a hierarchical latent modulated decoder. At the decoder, cross attention is utilized to exploit the contexts relevant to the given target followed by a stack of modulated MLP modules. 
The proposed VNP consists of a bottleneck encoder that efficiently encodes the context points into fewer informative context tokens,  refraining from high computational cost especially on complex signals. At the decoder side, cross-attention layers are utilized to exploit the contexts relevant to the given target followed by a stack of modulated MLP blocks. 
Particularly, we introduce multiple global latent variables to jointly modulate the MLP parameters, which facilitates the modeling of complex signal. 
% We have a few observations of the functions representing 1D, 2D or 3D signals. These context points, whether on-the-grid or off-the-grid, will be fed into the bottleneck transformer. The bottleneck transformer will transfer the context set into the context tokens, the number of which could be much less than the original number of context points. Then, queried by the target coordinates $X_T$, several cross-attention layers will produce the inputs of the subsequent MLPs. That is to say, the inputs of MLPs are adaptive features queried by each target coordinate, instead of solely the coordinates in previous INR works \citep{sitzmann2020implicit,tancik2020fourier,martin2021nerf}. 
% There are several modulated MLP blocks with hierarchical global latent variables $z_k = (z_1, z_2, ..., z_K)$. The parameters of the each MLP block can be modulated by these latent variables. The outputs of the final MLP block are the predicted signals at the corresponding target coordinates. It is noteworthy that the target signal value $Y_T$ will participate in the optimization to variationally learn the effective latent variables. During testing, the MLP parameters in the current block will be modulated by the latent variables generated from the output of the previous MLP block.

\subsection{Bottleneck Encoder in VNP} %Hierarchical Implicit Decoding}

Some NP methods encode the context points to produce local feature representations at target coordinates. The early work \citep{garnelo2018neural} uses simple MLP layers to learn features of the context set and suffers from underfitting problem. Attentive NP \citep{kim2019attentive} and Transformer NP \citep{nguyen2022transformer} enhances the feature representation by introducing point-wise self-attention with each context point taken as a token. However, when the number of context points is large (\egno in order to model an image signal with many details), the computational burden is heavy and such a framework is impractical. 

We address this issue by simply introducing a set tokenization module (set tokenizer) followed by self-attention layers. The set tokenizer is instantiated by a set convolution layer \citep{zaheer2017deep} which transforms the neighboring context points to a token. 
Taking a 2D image as an example, with the kernel size of $k \times k$ and stride of $k$, the set tokenizer can reduce the number of sample points by $k^2$ times (assume the image resolution is an integer multiple of $k$). By using the set tokenizer, the complexity of attention layers is reduced from the first term to the second term as below:
\begin{equation}
    \mathcal{O}(L_sN_C^2 + L_cN_CN_T)  \rightarrow  \mathcal{O}(L_s\frac{N_C^2}{k^4} + L_c\frac{N_C}{k^2}N_T),
\label{equation_complexity}
\end{equation}
where $L_s$ and $L_c$ are the number of the self-attention layers and the cross-attention layers, $N_C$ and $N_T$ are the number of context points and target points. 
The difference between the set convolution and the conventional convolution is that the former can handle missing points (\egno, in the application of inpainting) and the data that live "off the grid" (\egno~ time series data that observed irregularly at any time). 
The combination of set tokenizer and attention would empower our framework the flexibility in processing 3D signals, which is infeasible for ConvNP \citep{Gordon2020convolutional,foong2020meta} because ConvNP requires to preserve all the 3D grids, which is very inefficient in sparse 3D space. 

\begin{figure*}[t]
 \centering
 \begin{subfigure}{0.59\linewidth}
\includegraphics[scale=0.52, clip, trim=1.9cm 10.2cm 16cm 3cm]{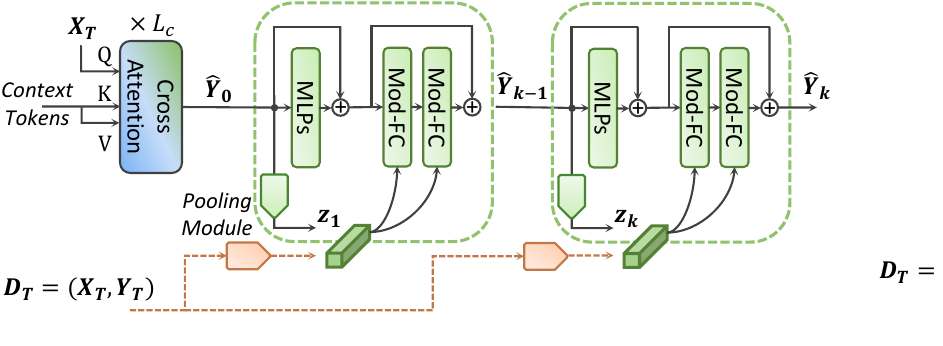}
 \caption{Pipeline of the decoder. \label{figure2a}}
 \end{subfigure}
 \begin{subfigure}{0.39\linewidth}
\includegraphics[scale=0.46, clip, trim=16.7cm 10cm 5.5cm 2.5cm]{./figures/figure2}
 \caption{Modulated MLP block. \label{figure2b}}
 \end{subfigure}
 \vspace{-0.2cm}
 \caption{Diagram of the decoder with the hierarchical latent modulated MLPs. \label{figure2}.}
\end{figure*}

%\subsection{Hierarchical Modulated MLPs}
\subsection{Hierarchical Latent Modulated MLPs}
\label{subsec:MLPs}

In the previous NP methods \citep{garnelo2018neural,kim2019attentive}, they learn a \emph{single} global latent Gaussian variable from the observed context set to model the distributions of a function. However, it potentially limits the expressiveness of the model. Intuitively, increasing the dimension of the latent variables may increase such flexibility, but in practice, this is not sufficient \citep{lee2020bootstrapping}. 
%We enhance the expressiveness of NPs by introducing a hierarchical implicit decoding architecture that exploit enhanced context for latent variable modelling. 

We design a hierarchical latent modulated decoder in order to better model the complex distribution of a function. We sequentially learn multiple global latent variables to modulate the parameters of MLP blocks. Figure \ref{figure2} shows the details of the designed hierarchical structure. % We propose a hierarchical implicit generation framework to gradually approach the ground truth function, conditioned on (partial) function observations. 

%Similar to ANP~\citep{}, 
The decoder consists of $L_c$ cross-attention blocks and $L_K$ modulated MLP blocks. With the target location $X_T$ as query and the context tokens from encoder as the keys and values, the cross-attention blocks output the target location features $\hat{Y}_0 \in \mathbb{R}^{T\times d}$, where $T$ is the number of target coordinates and $d$ is the feature dimension. The sequential modulated MLP blocks enable the exploitation of hierarchical global latent variables for the approximation of a complex signal. The output of the final ($L_{K}^{th}$) modulated MLP block goes through two MLP layers to estimate the probability of $Y_T$.

Figure \ref{figure2b} illustrates the detailed network structure of a modulated MLP block.  We build a Modulated MLP block by stacking two modulated MLP layers and two unmodulated MLP layers. This block refines  $\hat{Y}_{k-1}$ to output features $\hat{Y}_k$, which is the input of the next Modulated MLP block.
At the heart of each block is a latent variable $\mathbf{z} \in \mathbb{R}^{1\times d_\mathbf{z}}$ modeled by a Gaussian distribution.
%generated to refine the output of the previous cell $Y_{k-1}$. 
% The context condition $R_C$ and the target locations $\gamma(X_T)$ are known information, available to every decoding cell. 
The generation process of $\mathbf{z}_k$ (of the $k^{th}$ block) can be formulated as follows:
% Figure \ref{figure2b} illustrates the network structure of a modulated MLP block. 
% At the heart of each block is a low-dimensional latent variable $\mathbf{z} \in \mathbb{R}^{1\times d_z}$ modeled as Gaussian distribution.
% %generated to refine the output of the previous cell $Y_{k-1}$. 
% % The context condition $R_C$ and the target locations $\gamma(X_T)$ are known information, available to every decoding cell. 
% The generation process of $z_k$ (of the $k^{th}$ cell) can be formulated as follows:
\begin{equation}
\begin{aligned}
    %\rm R_{p}  = \rm MLP\ (concat\ [X_T, Y_{k-1}]) \in \mathbb{R}^{T\times d}, \\
    p_{\psi}(\mathbf{z}_k|\hat{Y}_{<k}, D_C, X_T) = \mathcal{N}(\mathbf{\mu_{\mathbf{z}_k}}, \mathbf{\sigma^2_{\mathbf{z}_k}}) & 	\leftarrow \rm{MLPs\ (AvgPool\ (MLPs} \it{(\hat{Y}_{k-1})))}, \\
    % q_{\phi}(z_k|\hat{Y}_{<k}, D_C, D_T) = \mathcal{N}(\mu_{z_k}, \sigma^2_{z_k}) & 	\leftarrow \rm{MLPs\ (AvgPool\ (MLPs} ([\hat{Y}_{k-1}, D_{T}]))).
\end{aligned}
\label{equation4}
\end{equation}
where MLPs refer to two MLP layers and an intermediate ReLU activation layer. The prediction results $\hat{Y}_{k-1}$ from the previous block (\ieno, the $(k-1)^{th}$ block) are used for calculating the conditional prior distribution of $\mathbf{z}_k$, \ieno, $p_{\psi}(\mathbf{z}_k|\hat{Y}_{<k}, D_C, X_T)$.
%, is modelled as Gaussian distribution, with mean and variance generated by another MLP based on ${\rm AvgPool\ (MLPs (Y_{k-1}))} \in \mathbb{R}^{1\times d}$. 
During training, the conditional posterior distribution $q_{\phi}(\mathbf{z}_k|\hat{Y}_{<k}, D_C, D_T)$ can be calculated as well, by incorporating the ground truth target signal values $Y_T$:
\begin{equation}
\begin{aligned}
    q_{\phi}(\mathbf{z}_k|\hat{Y}_{<k}, D_C, D_T) = \mathcal{N}(\mathbf{\mu_{\mathbf{z}_k}}, \mathbf{\sigma^2_{\mathbf{z}_k}}) & 	\leftarrow \rm{MLPs\ (AvgPool\ (MLPs} \it{([\hat{Y}_{k-1}, D_{T}])))}.
\end{aligned}
\label{equation4_extra}
\end{equation}
Marked with dashed lines, $Y_{T}$  only participates in training and cannot be accessed during inference. 
% As marked with dashed lines in our figures, $Y_{T}$  only participates in training ($\mathbf{z}$ samples from $q_{\phi}$) and cannot be accessed during inference ($\mathbf{z}$ samples from $p_{\psi}$). 
%In addition, when processing 3D scenes, in each block there is an additional rendering operator that will be introduced in detail in the section of 3D experiments.

%For a sampled realization of $z_k$, we use it to control the parameter weights of MLPs.
After sampling the low-dimensional latent variable $\mathbf{z}_k$, we use the modulated fully-connected (ModFC) layer \citep{karras2020analyzing,karras2021alias} to adjust the parameters of modulated MLP layers, taking a sampled realization of $\mathbf{z_k}$ as the style vector input. Unlike previous NP methods that concatenate the latent variable to every target coordinate, modulating the MLP parameters from the low-dimensional latent variables is a more flexible way to represent continuous functions \citep{sitzmann2020implicit}. Please see Appendix A for more details for the mechanism of the modulated MLP.
%We only modulate some of the MLPs to preserve the training stability.
% These modulated MLPs together with some unmodulated MLPs will enhance $\hat{Y}_{k-1}$,
% and output the refined function features $\hat{Y}_k$, which will be sent as the input of the next block. We only modulate some of the MLPs to preserve the training stability.
% The refining process of the target function values $Y_T$ adopts a residual connection:
% \begin{equation}
% \begin{aligned}
%     \rm Y_k' & = \rm  SelfAttention\ (R_{p}), \\
%     \rm Y_k & = \rm Y_k' + ModFC\ (Y'_{k}|z) \times2.
% \end{aligned}
% \label{equation7}
% \end{equation}

We model the function representations by multiple latent variables $(\mathbf{z}_1, \mathbf{z}_2, ..., \mathbf{z}_{L_K})$. The KL term in Eq. \ref{equation3}, which measures the mismatch between the approximated distribution and ground truth distribution, takes the hierarchical format as %Now we learn the function representations into disjoint groups, $z = (z_1, z_2, ..., z_L)$, similar to a top-down hierarchical VAE architecture \citep{sonderby2016ladder,vahdat2020nvae,rewon2021very}. Then the KL term in Eq. \ref{equation3}, which measures the mismatch between the approximated and ground truth functions, can be decomposed:
\begin{equation}
\begin{aligned}
    % D_{KL} = D_{KL} [q_{\phi}(z_1|D_t) || p_{\psi}(z_1|X_t, D_c)] + \sum_{i=2}^L D_{KL} [q_{\phi}(z_i|z_{<i}, D_t) || p_{\psi}(z_i|z_{<i}, X_t, D_c)].
    % D_{KL} = \sum_{k=1}^{L_K} \mathbb{E}_{q(\hat{Y}_{<k}|D_C, D_T)} [D_{KL} [q_{\phi}(\mathbf{z}_k|\hat{Y}_{<k}, D_C, D_T) || p_{\psi}(\mathbf{z}_k|\hat{Y}_{<k}, D_C, X_T)]],
    D_{KL} = \sum_{k=2}^{L_K} \mathbb{E}_{q_{\phi}(\mathbf{z}_{<k}| D_C, D_T)} [ &  D_{KL} [q_{\phi}(\mathbf{z}_k|\hat{Y}_{<k}, D_C, D_T) || p_{\psi}(\mathbf{z}_k|\hat{Y}_{<k}, D_C, X_T)]] \\
    & + D_{KL} [q_{\phi}(\mathbf{z}_1|D_C, D_T) || p_{\psi}(\mathbf{z}_1|D_C, X_T)],
\end{aligned}
\label{equation5}
\end{equation}
where $q_{\phi}(\mathbf{z}_{<k}|D_C, D_T) = \prod_{i=1}^{k-1} q_{\phi}(\mathbf{z}_{i}|\hat{Y}_{<i}, D_C, D_T)$ is the approximate posterior of $\mathbf{z}_{<k}$.
%up to the $(k-1)^{th}$ latent variable.
This decomposed KL term and the reconstruction term formulate the objective:
\begin{equation}
    \mathcal{L} =  \mathbb{E}_{\mathbf{z}_{1:L_K} \sim q_{\phi}(\mathbf{z}_{1:L_K}|D_T)} [- \log p_{\theta}(Y_T |  \mathbf{z}_{1:L_K}, X_T, D_C)] + \beta \cdot D_{KL},
\label{equation6}
\end{equation}
where $\beta$ denotes a weight to balance the importance between the two terms. In our experiments on 2D and 3D signals, we multiply the KL term with a small weight $\beta$ to better capture the uncertainty of function distributions \citep{irina2017beta}.
% in Eq. \ref{equation6}. In our experiments on 2D and 3D signals, we multiply the KL term with a small weight $\beta$ to better capture the uncertainty of function distributions \citep{irina2017beta}.
% \begin{equation}
%     \mathcal{L} =  \mathbb{E}_{\mathbf{z}_{1:L_K} \sim q_{\phi}(\mathbf{z}_{1:L_K}|D_T)} [- \log p_{\theta}(Y_T |  \mathbf{z}_{1:L_K}, X_T, D_C)] + \beta \cdot D_{KL}
% \label{equation6}
% \end{equation}
In addition, we emphasize that although there are some prior works designing hierarchical architecture for VAEs \citep{sonderby2016ladder,vahdat2020nvae,rewon2021very} or double latent variable models for NP \citep{wang2020doubly}, our proposed hierarchical architecture is different in principle. Here, every latent variable is a low-dimensional vector obtained after average pooling. Therefore, the designed hierarchical architecture can deal with arbitrary target coordinates and thereby can be used for boosting the performance of approximating the global structure of continuous functions.

\section{Experiments}

The proposed Versatile Neural Process (VNP), as an efficient meta-learner of implicit neural representations, can be implemented into a variety of tasks. We evaluate the effectiveness of VNP on 1D function regression (subsection \ref{subsec-1D}), 2D image completion and superresolution (subsection \ref{subsec-2D}), and view synthesis for 3D scenes (subsection \ref{subsec-3D}), respectively. 

\vspace{-0.2cm}
\subsection{1D Signal Regression} \label{subsec-1D}

We implement the proposed VNP to learn the implicit neural representations for 1D signal regression. 
%By representing the signals as continuous functions, we face an 1D function regression task that has long been studied by the Neural Process family \citep{garnelo2018neural,garnelo2018conditional}. 
This classical 1D regression aims at predicting the function values at given target locations, conditioned on several observations of the samples from the function. 
Following \citep{kim2019attentive}, we measure the performance by considering the context set likelihood and target set likelihood, which reflects the context reconstruction error and target prediction error, respectively. 
% By representing the signals as continuous functions, we face an 1D function regression task that has long been studied by the Neural Process family \citep{garnelo2018neural,garnelo2018conditional}. 
% This classical 1D regression aims at predicting the function values at target locations, with several observations of the samples from the function. 
% Following \citep{kim2019attentive}, we measure the performance by considering the context likelihood and target likelihood, which reflect the context reconstruction error and target prediction error, respectively. 

\textbf{Settings.} We train the models on synthetic functions drawn from prior function distributions synthesized with different kernels (RBF, Matern) by following \citep{Gordon2020convolutional,kim2022neural}. 
For the  evaluated methods, we employ importance weighted sampling \citep{yuri2016importance} to evaluate the log likelihood, where the last four methods in Table \ref{table1} are measured by sampling latent variables from the posterior distribution and calculating the tighter ELBO with importance weighted sampling. 
%Our hierarchical models consist of six decoding cells in this 1D function regression task. 
%For fair comparison, we build our baseline with the similar size of network. 
For fair comparison, we manage to keep the network size comparable with that of other methods, which is completed by adjusting channel number. Please refer to Appendix B for detailed settings.
%For fair comparison, we manage to keep network capacity around 2M, almost the same as the baseline models
% \textbf{Settings.} We train the models on synthetic functions drawn from prior function distributions synthesized with different kernels (RBF, periodic, Matern) by following \citep{Gordon2020convolutional,kim2022neural}. 
% For all the evaluated methods, we employ importance weighted sampling \citep{yuri2016importance} to evaluate the log likelihood. 
% %Our hierarchical models consist of six decoding cells in this 1D function regression task. 
% %For fair comparison, we build our baseline with the similar size of network. 
% Note although our hierarchical models consist of multiple blocks, we manage to keep network capacity around 2M, almost the same as the baseline models by reducing the channel number in VNP. Please refer to Appendix A for detailed experimental settings.

\textbf{Comparison with the State-of-the-Arts}. We compare our method with conditional neural process (CNP) \citep{garnelo2018conditional}, (stacked) attentive neural process (ANP) \citep{kim2019attentive} (with stacked self-attention layers), bootstrapping attentive neural process (BANP) \citep{lee2020bootstrapping}, and convolutional neural process (ConvNP) \citep{foong2020meta}. Table \ref{table1} shows the results in terms of log likelihood.
Most of the models can provide satisfactory reconstruction results for context points, except CNP which suffers from underfitting problem \citep{garnelo2018neural}. On the context points, ours also achieves comparable performance. However, the prediction results of other methods on the unseen target points are much worse than that on the context sets. In contrast, our final model, VNP, outperforms previous approaches by a large margin at the target points on both of the function distributions. Note that although BANP attempts to increase the expressiveness of function representations by using latent variable bootstrapping from the perspective data resampling.% Ours outperforms BANP significantly.

% \textbf{Results}. Table \ref{table1} summarizes the numerical results in this regression task. 
% We compare our method with conditional neural process (CNP) \citep{garnelo2018conditional}, (stacked) attentive neural process (ANP), bootstrapping attentive neural process (BANP) \citep{lee2020bootstrapping} and convolutional neural process (ConvNP) \citep{foong2020meta}. 
% Most of the models can provide satisfactory reconstruction results at context points, except CNP which approximates the whole function with a simple global-pooling feature \citep{garnelo2018neural}. However, all previous methods including ANP, BANP and ConvNP may overfit for context points while underfitting for target points. 
% Note that although BANP attempts to increase the expressiveness of function representations with latent variable bootstrapping, our experiments find the enhanced attentive neural process (ANP) with multiple stacked attention layers would achieve better performance than BANP, but still have a large gap with our method.
% Our final model, Transformer-based Hierarchical Neural Processes (THNP), outperforms previous work by a large margin at target points on all three function distributions.
% Nevertheless, our proposed HINP always increases the expressiveness of function representations under the same network capacity, which is shown to outperform both the ANP and BANP on all three function distributions.

\begin{table*}[t]
\centering
\small
\renewcommand{\arraystretch}{1.1}
\resizebox{0.9\textwidth}{!}{
\begin{tabular}{cccccc}
\hline
\multicolumn{1}{l}{\multirow{2}{*}{}} & \multicolumn{2}{c}{RBF kernel GP}                        & \multicolumn{2}{c}{Matern kernel GP}      & \multicolumn{1}{c}{\multirow{2}{*}{Parameters}}              \\ \cline{2-5} 
\multicolumn{1}{l}{}                  & \multicolumn{1}{c}{context} & \multicolumn{1}{c}{target} & \multicolumn{1}{c}{context} & \multicolumn{1}{c}{target}  & \multicolumn{1}{c}{} \\ \hline
CNP           & 1.023$\pm$0.033                 & 0.019$\pm$0.015         & 0.935$\pm$0.036                 & -0.124$\pm$0.010               & 0.99 M   \\ \hline
BANP                                 & 1.380$\pm$0.000                 & 0.267$\pm$0.001                & 1.380$\pm$0.002                 &       0.072$\pm$0.002    & 1.58 M   \\ \hline
ConvNP                                  & 1.382$\pm$0.001                 & 0.275$\pm$0.001         & 1.383$\pm$0.001                 & 0.081$\pm$0.008        & 1.97 M        \\ \hline
% Stacked ANP (old metric)                                  & 1.381$\pm$0.001                 & 0.292$\pm$0.002         & 1.381$\pm$0.001                 & 0.082$\pm$0.003   & 1.52 M  \\ \hline
Stacked ANP                                  & 1.381$\pm$0.001                 & 0.400$\pm$0.004         & 1.381$\pm$0.001                 & 0.183$\pm$0.012   & 1.52 M  \\ \hline
Stacked ANP +                                  & 1.381$\pm$0.001                 & 0.406$\pm$0.006         & 1.381$\pm$0.001                 & 0.188$\pm$0.014   & 2.31 M  \\ \hline
HNP                                  & 1.374$\pm$0.002                 & 0.561$\pm$0.003                & 1.377$\pm$0.001       &   0.336$\pm$0.008          & 1.66 M        \\ 
HNP-Mod                                & 1.379$\pm$0.001                 & 0.627$\pm$0.020                & 1.371$\pm$0.004       &   0.370$\pm$0.023       & 1.96 M         \\ 
VNP                 & 1.377$\pm$0.004                 & \textbf{0.651$\pm$0.001}                &     1.376$\pm$0.004      &   \textbf{0.439$\pm$0.007}  & 2.29 M   \\ \hline

% \multicolumn{1}{l}{HAIG+FF}           &      1.375$\pm$0.000          &      \textbf{0.962$\pm$0.000}minipage       &         1.352$\pm$0.000            &        0.704$\pm$0.000           &      1.194$\pm$0.000       &      -0.176$\pm$0.000    \\ \hline
\end{tabular}}
\caption{The test log likelihood (larger is better) on the synthetic 1D regression experiment. The proposed VNP outperforms previous methods by a large margin. Both the pre-processing transformer and the hierarchical structure improve the expressiveness of function representations. Stacked ANP + means the enhanced version of Stacked ANP with more channels.  \label{table1}}
\vspace{-0.3cm}
\end{table*}

\textbf{Ablation Study.} 
We conduct a group of ablation study on this 1D regression task to investigate the effectiveness of different components. Based on ANP, we first build a Hierarchical Neural Process (HNP) with hierarchical latent variables. Note that similar to ANP, the implemented HNP still concatenates every target coordinate with the global latent variables. We observe significant improvements in Table \ref{table1} by comparing HNP with ANP, demonstrating that the hierarchical global latent variables boost the performance of function approximation.
Secondly, the modulated MLP layer for implicit function parameterization can be further equipped in HNP, referred to as HNP-Mod. Compared with the concatenation of latent variable, using low-dimensional latent variables to modulate the network parameters enables more flexible function approximating. Based on HNP-Mod, we finally add the bottleneck encoder to learn conditional network inputs that are adaptive to target locations, building our final model VNP. 
%Specifically, the context points are pre-processing into grid points with set convolution, which are then tokenized with downsampling convolution. We use these tokens
In Table \ref{table1}, we can find this powerful pre-processing encoder can further improve the performance, because the set tokenizer (set convolution here) can preserve locality to learn more appropriate features. 

In Table \ref{table_ablation}, we further ablate on the detailed hierarchical structures in our model. As we can observe, when we increase the number of Modulated MLP blocks, the prediction performance (at target points) is improved until saturated with 6 blocks (\ieno, $L_K=6$). 
%If we further increase the block number, the training period will be unstable and results in larger variance. 
In addition, the column with "6 / single $\mathbf{z}$" means we use exactly the same network structures as $L_K=6$, but instead only impose the KL constraint to the final latent variable $\mathbf{z}_6$. It is found that the performance will also drop dramatically, which verifies that the gains come from the hierarchical design instead of the increased network capacity.
This ablation study guides us to set the number of Modulated MLP blocks as 6 to compare with other methods, which can keep a balance between the performance and the complexity.

\begin{table}[]
\centering
\resizebox{0.98\textwidth}{!}{
\renewcommand{\arraystretch}{1.1}
\begin{tabular}{ccccccc}
\hline
\multirow{2}{*}{} & \multicolumn{6}{c}{number of Modulated MLP blocks} \\ \cline{2-7} 
                  & 0           & 2       & 4       & 6      & 8   & 6 / single $\mathbf{z}$   \\ \hline
context           & 1.375$\pm$0.001 &   1.376$\pm$0.001  &  1.376$\pm$0.001  & 1.376$\pm$0.001  &  1.376$\pm$0.001 & 1.376$\pm$0.001 \\ \hline
target            & 0.076$\pm$0.001 &   0.336$\pm$0.028  &  0.371$\pm$0.001  & \textbf{0.439$\pm$0.007} &  0.435$\pm$0.028 &  0.245$\pm$0.003 \\ \hline
\end{tabular}}
\vspace{-0.2cm}
\caption{Ablation study about the detailed hierarchical structure on Matern kernel. Here we also provide the results of test log-likelihood (larger is better). \label{table_ablation}}
\vspace{-0.1cm}
\end{table}

\begin{figure*}[t]
 \centering
 \begin{subfigure}{0.49\linewidth}
\includegraphics[scale=0.28, clip, trim=3.8cm 10cm 2.2cm 0.4cm]{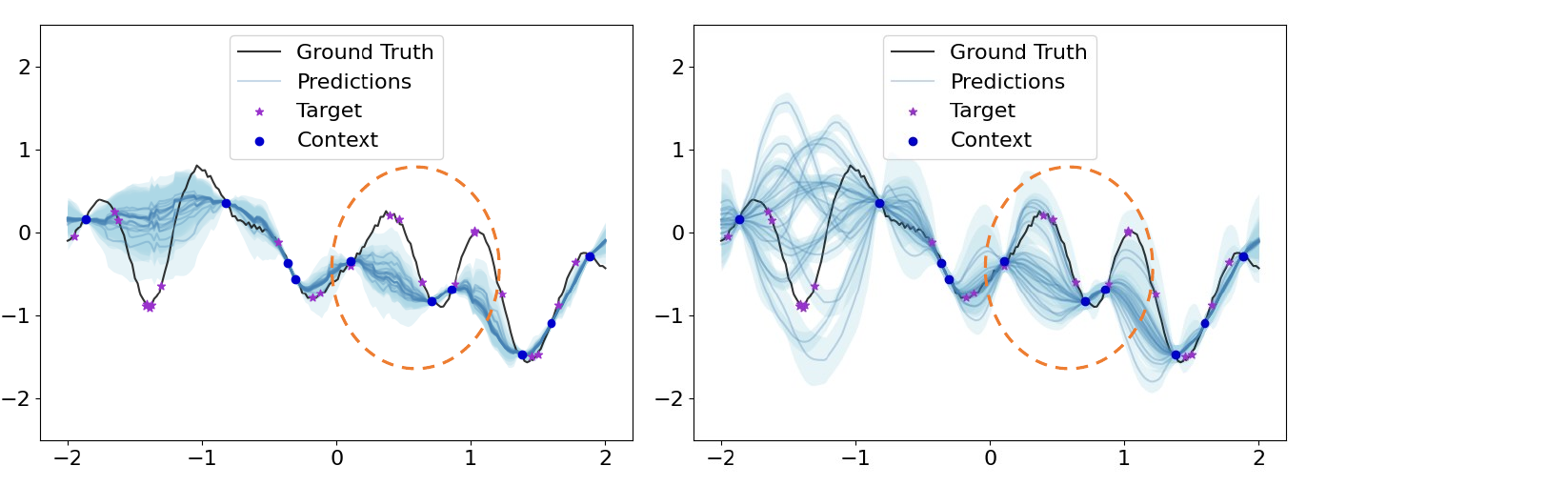}
 \caption{Matern kernel. ANP vs. our VNP.\label{figure4b}}
 \end{subfigure}
 \begin{subfigure}{0.49\linewidth}
\includegraphics[scale=0.28, clip, trim=2.8cm 10cm 2.5cm 0.7cm]{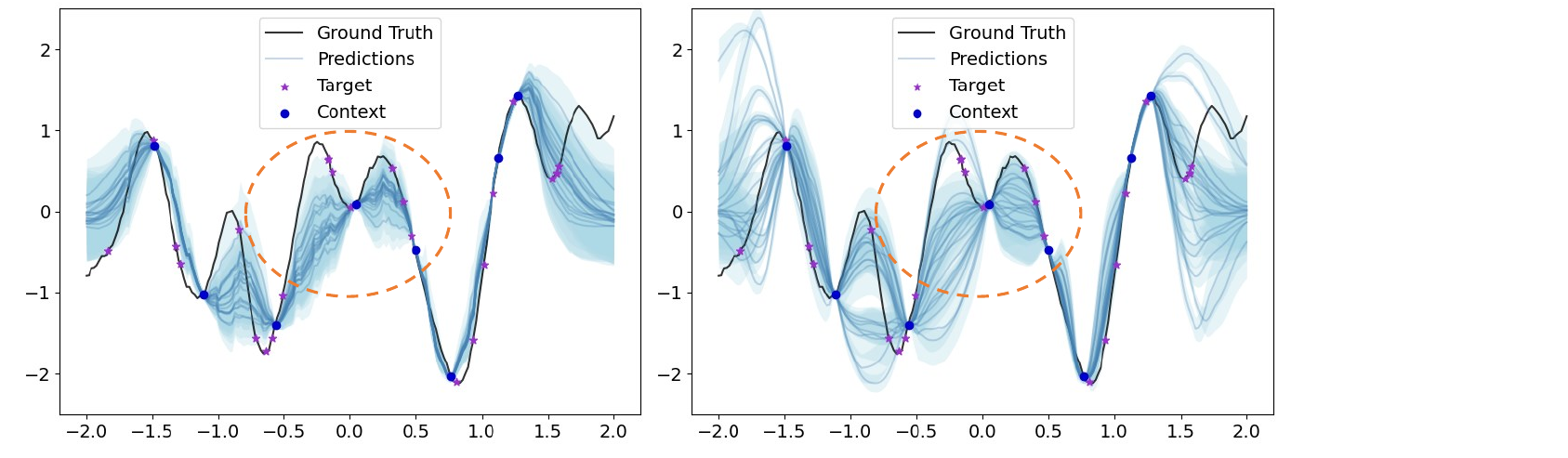}
 \caption{RBF kernel. ANP vs. our VNP.\label{figure4a}}
 \end{subfigure}
 \vspace{-0.2cm}
 \caption{Visualizations of the 1D regression results. VNP delivers diverse function prediction results, while ANP \citep{kim2019attentive} tends to underestimate the function variances at some target locations. }
\vspace{-0.1cm}
\label{figure3}
\end{figure*}

\vspace{-0.1cm}

%\tcr{Zongyu, please re-organize this section.  Ablation first, comparision with SOTA second, visilization finally.}

\textbf{Visualizations.} We visualize the obtained function distributions from stacked ANP \citep{kim2019attentive} and our VNP, by sampling the latent variables 20 times. Given partial observations of a signal, there exists uncertainty on the continuous function since there are many possible ways to interpret these observations (i.e., context set). An excellent Neural Process model should be able to model such uncertainty through the fitted functions. In other words, the generated functions conditioned on the context set should approximate the data distribution and cover the target set points. As shown in Figure \ref{figure3}, we can see that the generated functions from the stacked ANP cannot predict the target points accurately, \egno, those regions marked by orange. In contrast, our method provide good approximations for the continuous functions, where the distribution covers the groundtruth function.

\begin{figure*}[t]
\centering
\begin{minipage}{0.59\linewidth}
\begin{subfigure}{\linewidth}
\includegraphics[scale=0.57, clip, trim=8.2cm 5.2cm 12cm 7cm]{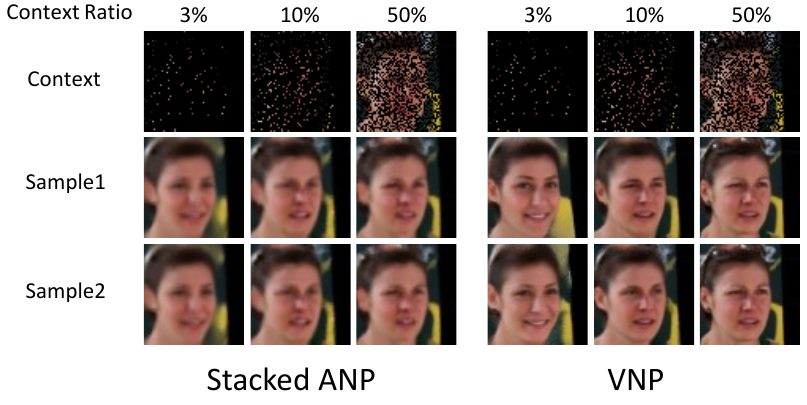}
\end{subfigure}
\caption{Qualitative comparisons with Stacked ANP \citep{kim2019attentive}. Our VNP presents diverse and realistic image completion results.\label{figure5}}
\end{minipage}
\hspace{0.005\linewidth}
% \begin{minipage}{0.34\linewidth}
% \renewcommand{\arraystretch}{1.5}
% \resizebox{\linewidth}{!}{
% \begin{tabular}{|c|c|}
% \hline
%   CelebA64     & NLL (lower is better)                                     \\ \hline
% Stacked ANP & 2.88 \\ \hline
% VNP (block=2) & 2.83 \\ \hline
% VNP (block=4) & 2.80 \\ \hline
% VNP (block=6) & \textbf{2.78} \\ \hline
% \end{tabular}}
% \end{minipage}
\begin{minipage}{0.39\linewidth}
\renewcommand{\arraystretch}{1.5}
\resizebox{\linewidth}{!}{
\begin{tabular}{|c|c|}
\hline
  CelebA64     & \makecell[c]{NLL (lower is better) \\ context ratio = 0.03}  \\ 
\hline 
Stacked ANP & 2.988 \\ 
\hline
\makecell[c]{VNP, $ks$ = 1, $L_K$ = 1}  & 2.994 \\ 
\hline
\makecell[c]{VNP, $ks$ = 1, $L_K$ = 6}  & 2.953 \\ 
\hline
VNP, $ks$ = 2, $L_K$ = 6 & 2.952 \\ 
\hline
VNP, $ks$ = 4, $L_K$ = 6 & 2.964 \\ 
\hline
\end{tabular}}
\captionof{table}{Quantitative results measured by Eq.\ref{equation6} on the test set. Lower is better. $ks$ is the kernel size in set tokenizer. $L_K$ is the number of Modulated MLP blocks. \label{tab:kernel}}
\end{minipage}
% \vspace{-0.1cm}
% \caption{Left: Qualitative comparisons with Stacked ANP \citep{kim2019attentive}. Our VNP presents diverse and realistic image completion results. \tcb{Right: Quantitative results measured by Eq.\ref{equation6} on the test set. Lower is better. $k$ is the kernel size in set tokenizer, and $L_K$ is the hierarchical block number.} \label{figure5}}
\vspace{-0.1cm}
\end{figure*}

\begin{figure*}[t]
 \centering
 \begin{subfigure}{0.29\linewidth}
\includegraphics[scale=0.56, clip, trim=7cm 10cm 19.4cm 4.5cm]{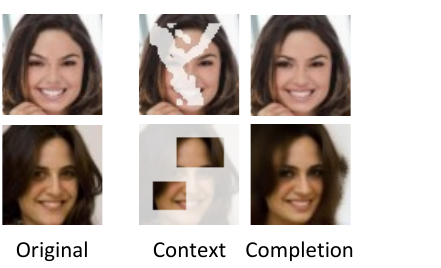}
 \caption{\label{figure4a}}
 \end{subfigure}
 \begin{subfigure}{0.36\linewidth}
\includegraphics[scale=0.56, clip, trim=14.7cm 10cm 9.6cm 4.5cm]{./figures/figure4}
 \caption{\label{figure4b}}
 \end{subfigure}
 \begin{subfigure}{0.33\linewidth}
\includegraphics[scale=0.56, clip, trim=24.1cm 10cm 2.2cm 4.3cm]{./figures/figure4}
 \caption{\label{figure4c}}
 \end{subfigure}
\vspace{-0.2cm}
 \caption{Visualization of VNP results on CelebA64 \citep{liu2015deep}. (a) Image Completion. (b) Superresolution to arbitrary size. (c) Handling image completion and superresolution simultaneously.\label{figure4}}
\end{figure*}

\subsection{2D Images Completion and Superresolution} \label{subsec-2D}

A 2D image can be modeled by a continuous function that maps the 2D pixel coordinates to the color values. We implement our VNP framework for learning the continuous function of 2D images, which supports tasks such as image completion and super-resolution to arbitrary size. 

\textbf{Settings.} We conduct experiments on CelebA dataset \citep{liu2015deep}, mainly with the resized resolution of $64\times64$.
%We conduct experiments on CelebA64 and Celeba178 datasets, which are obtained by resizing CelebA dataset \citep{liu2015deep} into $64\times64$ and $178\times178$ resolution respectively.
%different resolutions, including CelebA64 and Celeba178. 
Due to the limited representation capability or the high requirement on computation resources, most previous NP methods \citep{kim2019attentive,lee2020bootstrapping} are in general trained and evaluated on relatively low resolution such as $32\times 32$. 
Our framework enables the processing of complex signal with higher resolution. 
We train a single model that supports image completion and super-resolution tasks. During training, we control the context ratio as 0.03, which means 3\% pixels are taken as context set. The target ratio is set as 0.15 %The target ratio is set as 0.25 on CelebA64 and 0.15 on CelebA178. 
and the target set has partial intersection with the context set. More detailed experimental settings can be found in Appendix B.

\textbf{Visualizations}. We compare our VNP with stacked ANP \citep{kim2019attentive} tested with three different context ratios respectively as shown in Figure \ref{figure5}. Our VNP generates diverse and realistic image completion results with fine details. In comparison, the results of Stacked ANP are blurred. 

In Figure \ref{figure4}, we show that the proposed VNP achieves satisfactory results for image completion and superresolution on CelebA64 dataset. As a unified framework, with a single model, it can support image completion, superresolution, and the concurrent of completion and superresolution. 

\textbf{Influence of Patch/Kernel Size}. We conduct an ablation study to investigate the influence of the patch/kernel size in our set tokenizer. We set the stride the same as the kernel size. The results are shown in Table~\ref{tab:kernel}. It is observed that using larger kernel size does not degrade the performance obviously. In addition, using hierarchical structure brings gains.
%We can see that using set tokenizer will not hurt the performance obviously, while it can reduce the computational complexity when the kernel size is larger than 1.}
%Due to the limit of GPU memory, we finally set the number of hierarchical blocks also as 6 for this 2D experiment.
 
%Since the bottleneck encoder has the advantage of reducing the computational complexity, our method is applicable to images with resolution of $178\times178$. 

\begin{table}[!t]
\centering
\renewcommand{\arraystretch}{1.0}
\begin{tabular}{c|ccc|ccc}
\hline
\multirow{3}{*}{GFLOPs} & \multicolumn{3}{c}{CelebA64}      & \multicolumn{3}{c}{CelebA178}     \\ \cline{2-7} 
                        & \multicolumn{3}{c}{Context Ratio} & \multicolumn{3}{c}{Context Ratio} \\ \cline{2-7} 
                        & 0.05      & 0.25      & 1.0       & 0.05       & 0.25      & 1.0      \\ \hline
Stacked ANP             & 17.1      & 39.8      & 198.5     & 190.0      & 867.2    & \tcgr{12150}      \\ \hline
Our VNP                 & 27.6      & 27.6      & 27.6      & 204.5      & 204.5     & 204.5    \\ \hline
\end{tabular}
\caption{Comparing our VNP with stacked ANP \citep{kim2019attentive} about the complexity, measure by FLOPs. The bottleneck encoder plays an important role in reducing the computational cost. The statistic in gray is estimated since we have an upper bound limitation of GPU memory. \label{table2}}
\vspace{-0.2cm}
\end{table}

\textbf{Complexity Comparisons.} Thanks to our bottleneck encoder design, the proposed VNP is a practical framework for complex signal modeling, where the number of context points as condition is usually large. For comparison, we calculate the GFLOPs of stacked ANP and our VNP when target ratio is 0.25 on CelebA64 and CelebA178 respectively. Here, the kernel size of set tokenizer is 4 on CelebA64 and 10 on CelebA178. 
%Note both these two models are designed with the same number of stacked self-attention and cross-attention layers. 
The computational complexity comparisons in terms of GFLOPs are shown in Table \ref{table2}. Since we use set tokenizer to reduce the number of tokens and process the context set in image grids, the GFLOPs of VNP would be smaller than that of Stacked ANP, especially when the context ratio is high.

\begin{figure*}[t]
 \begin{subfigure}{0.49\linewidth}
 \centering
\includegraphics[scale=0.47, clip, trim=2.8cm 7.9cm 13cm 6.4cm]{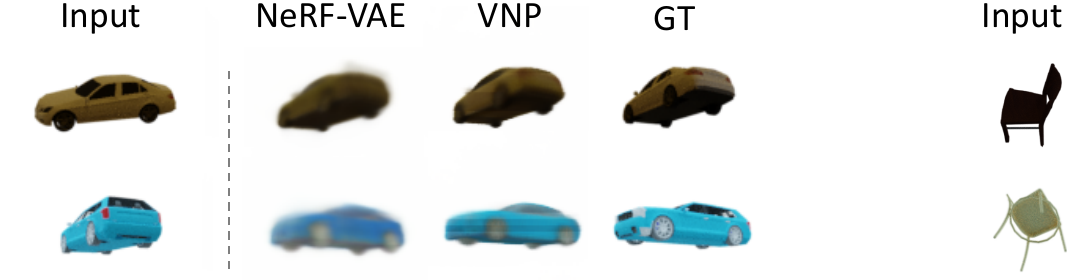}
\caption{Cars. \label{figure8a}}
 \end{subfigure}
 \begin{subfigure}{0.49\linewidth}
 \centering
\includegraphics[scale=0.47, clip, trim=17cm 7.9cm 3cm 6.4cm]{./figures/figure7}
\caption{Chairs. \label{figure8b}}
 \end{subfigure}
 \vspace{-0.2cm}
\caption{Novel view synthesis on ShapeNet \citep{chang2015shapenet} Objects. Our VNP presents more realistic prediction results than the blurry results of NeRF-VAE \citep{kosiorek2021nerf}. \label{figure8}}
\end{figure*}

\begin{table}[t]
\centering
\renewcommand{\arraystretch}{1.1}
\begin{tabular}{ccccc}
\hline
\multicolumn{2}{c}{\multirow{2}{*}{}}                                  & \multicolumn{3}{c}{PSNR (dB)} \\ \cline{3-5} 
\multicolumn{2}{c}{}                                                   & Cars       & Lamps  & Chairs     \\ \hline
\multirow{2}{*}{Deterministic} & Learned Init \citep{chen2021learning}  &    22.80      & 22.35   &  18.85  \\ \cline{2-5} 
                               & Trans-INR \citep{chen2022transformers}    &   23.78    &  22.76   &  \textbf{19.66} \\ \hline
\multirow{2}{*}{Probabilistic} & NeRF-VAE \citep{kosiorek2021nerf}     &  21.79   &  21.58   &  17.15 \\ \cline{2-5} 
                               & Our VNP                               &    \textbf{24.21}   &   \textbf{24.10}    &  19.54 \\ \hline
\end{tabular}
\caption{The quantitative results of one-shot novel view synthesis. We compare the proposed VNP with both deterministic and probabilistic methods. All of these methods are able to produce INRs representing the 3D scenes within few steps. \label{table3}}
\end{table}

%We implement our VNP for 3D scene regression. Conditioned on a few observed views, we learn the function of the 3D signal that allows us to generate any novel view (\ieno, view synthesis).

%Most previous works \citep{mildenhall2020nerf,martin2021nerf} requires expensive optimization to fit the neural network to represent a scene. We focus on the task of one-shot novel view synthesis in this section to evaluate our proposed VNP. 
\subsection{View synthesis on 3D Scene} \label{subsec-3D}

Implicit neural representations excel at representing 3D signals. However, since 3D signals are usually much more complex, most previous works \citep{mildenhall2020nerf,martin2021nerf} require expensive optimization to fit the neural networks to represent the scenes. We focus on the task of view synthesis in this section to evaluate our proposed VNP. 

\textbf{Settings.} We follow the spirits of NeRF \citep{mildenhall2020nerf} that fits a network to map the world coordinate into the corresponding RGB values and volume density. We use bottleneck encoder (tokenization from the image patches) to calculate the adaptive input of the MLPs, queried by target world coordinates. Then the hierarchically learned global latent variable will modulate the parameters of MLPs to predict the RGB values and volume density. Note that there is an extra volume rendering process inside each decoding block before the pooling module, because it requires transferring the world coordinate to the image coordinate to compute the latent distribution of $\mathbf{z}_k$. We conduct experiments on ShapeNet \citep{chang2015shapenet} objects, including three sub-datasets: cars, lamps, and chairs. More details on the network structures and hyper parameters can be found in Appendix B.
% \textbf{Settings.} We follow the spirits of NeRF \citep{mildenhall2020nerf} that fits MLPs to map the world coordinate into the corresponding RGB values and volume density. However, we will first use bottleneck transformer (tokenization from the image patches) to calculate the adaptive input of the MLPs, queried by target world coordinate. Then the hierarchically learned global latent variable will modulate the parameters of MLPs to predict the RGB values and volume density in a single forward pass. Note that as mentioned in Section 3, there is an extra volume rendering process inside each decoding block because it requires transferring the world coordinate to the image coordinate to compute the latent distribution of $z_k$. We conduct this one-shot view synthesis experiments with ShapeNet \citep{chang2015shapenet} objects, including three sub-dataset: cars, lamps and chairs. The specific model structures and other hyper parameters can be found in Appendix A.
%Other specific experimental settings are shown in Appendix C.

\textbf{Comparisons.} Our VNP model can generate the implicit neural representations \wrt the previously unseen scene in a single forward pass. In this 3D task, one prior work NeRF-VAE \citep{kosiorek2021nerf} can also \textit{generate} the scene from the randomly sampled latent variable. We make quantitative and qualitative comparisons with NeRF-VAE. In addition, we compare with \citep{chen2021learning} and \citep{chen2022transformers} which can also learn implicit neural representations of the previously unseen scene, although they are not in the family of probabilistic models and thus cannot model function distributions. 
As shown in Table \ref{table3},
%\footnote{Updated to align with the checkpoints on \url{https://github.com/ZongyuGuo/Versatile-NP}}, 
our method achieves the best performance in terms of Peak Signal-to-Noise Ratio (PSNR) on Cars and Lamps, and is comparable with Trans-INR \citep{chen2022transformers} on Chairs. The visualizations in Figure \ref{figure8a} show that VNP produces much better novel-view prediction results than NeRF-VAE, which is also a probabilistic generative model. 
%Note that during optimization, VAE estimates the distribution of the whole dataset. In contrast, NPs estimate the conditional distribution of the target points given some context points, which is more adaptive to a specific signal.

\section{Conclusion and Discussion}

The Neural Processes family provides an efficient way for learning implicit neural representations by approximating the function distribution when only partial observation of a signal is given.
We propose an efficient NP framework, Versatile Neural Processes (VNP), that largely increases the capability of approximating functions. 
%Specifically, we have a bottleneck encoder that produces \tcr{fewer informative} context tokens, relieving the high computation requirement especially when the number of context input-output pairs are huge.
Our bottleneck encoder and hierarchical latent modulated decoder enables strong modeling capability to the complex signals.  
Through comprehensive experiments, we show the effectiveness of the proposed VNP on 1D, 2D and 3D signals. Our work demonstrates the potential of neural process as a promising solution for efficient learning of INRs in complex 3D scenes.

\subsubsection*{Acknowledgments}

This work was supported in part by NSFC under Grant U1908209, 62021001.

\medskip

\bibliography{iclr2023_conference}

\begin{thebibliography}{46}
\providecommand{\natexlab}[1]{#1}
\providecommand{\url}[1]{\texttt{#1}}
\expandafter\ifx\csname urlstyle\endcsname\relax
  \providecommand{\doi}[1]{doi: #1}\else
  \providecommand{\doi}{doi: \begingroup \urlstyle{rm}\Url}\fi

\bibitem[Burda et~al.(2016)Burda, Grosse, and
  Salakhutdinov]{yuri2016importance}
Yuri Burda, Roger~B. Grosse, and Ruslan Salakhutdinov.
\newblock Importance weighted autoencoders.
\newblock In \emph{4th International Conference on Learning Representations,
  {ICLR} 2016}, 2016.

\bibitem[Chang et~al.(2015)Chang, Funkhouser, Guibas, Hanrahan, Huang, Li,
  Savarese, Savva, Song, Su, et~al.]{chang2015shapenet}
Angel~X Chang, Thomas Funkhouser, Leonidas Guibas, Pat Hanrahan, Qixing Huang,
  Zimo Li, Silvio Savarese, Manolis Savva, Shuran Song, Hao Su, et~al.
\newblock Shapenet: An information-rich 3d model repository.
\newblock \emph{arXiv preprint arXiv:1512.03012}, 2015.

\bibitem[Chen et~al.(2021{\natexlab{a}})Chen, He, Wang, Ren, Lim, and
  Shrivastava]{chen2021nerv}
Hao Chen, Bo~He, Hanyu Wang, Yixuan Ren, Ser~Nam Lim, and Abhinav Shrivastava.
\newblock Nerv: Neural representations for videos.
\newblock volume~34, 2021{\natexlab{a}}.

\bibitem[Chen \& Wang(2022)Chen and Wang]{chen2022transformers}
Yinbo Chen and Xiaolong Wang.
\newblock Transformers as meta-learners for implicit neural representations.
\newblock \emph{arXiv preprint arXiv:2208.02801}, 2022.

\bibitem[Chen et~al.(2021{\natexlab{b}})Chen, Liu, and Wang]{chen2021learning}
Yinbo Chen, Sifei Liu, and Xiaolong Wang.
\newblock Learning continuous image representation with local implicit image
  function.
\newblock In \emph{Proceedings of the IEEE/CVF Conference on Computer Vision
  and Pattern Recognition}, pp.\  8628--8638, 2021{\natexlab{b}}.

\bibitem[Chen \& Zhang(2019)Chen and Zhang]{chen2019learning}
Zhiqin Chen and Hao Zhang.
\newblock Learning implicit fields for generative shape modeling.
\newblock In \emph{Proceedings of the IEEE/CVF Conference on Computer Vision
  and Pattern Recognition}, pp.\  5939--5948, 2019.

\bibitem[Child(2021)]{rewon2021very}
Rewon Child.
\newblock Very deep vaes generalize autoregressive models and can outperform
  them on images.
\newblock In \emph{9th International Conference on Learning Representations,
  {ICLR} 2021}, 2021.

\bibitem[Dupont et~al.(2021)Dupont, Golinski, Alizadeh, Teh, and
  Doucet]{dupont2021coin}
Emilien Dupont, Adam Golinski, Milad Alizadeh, Yee~Whye Teh, and Arnaud Doucet.
\newblock {COIN}: {CO}mpression with implicit neural representations.
\newblock In \emph{Neural Compression: From Information Theory to Applications
  -- Workshop @ ICLR 2021}, 2021.

\bibitem[Dupont et~al.(2022)Dupont, Loya, Alizadeh, Golinski, Teh, and
  Doucet]{dupont2022coin++}
Emilien Dupont, Hrushikesh Loya, Milad Alizadeh, Adam Golinski, Y~Whye Teh, and
  Arnaud Doucet.
\newblock Coin++: Neural compression across modalities.
\newblock \emph{Transactions on Machine Learning Research}, 2022\penalty0 (11),
  2022.

\bibitem[Finn et~al.(2017)Finn, Abbeel, and Levine]{finn2017model}
Chelsea Finn, Pieter Abbeel, and Sergey Levine.
\newblock Model-agnostic meta-learning for fast adaptation of deep networks.
\newblock In \emph{International conference on machine learning}, pp.\
  1126--1135. PMLR, 2017.

\bibitem[Foong et~al.(2020)Foong, Bruinsma, Gordon, Dubois, Requeima, and
  Turner]{foong2020meta}
Andrew Foong, Wessel Bruinsma, Jonathan Gordon, Yann Dubois, James Requeima,
  and Richard Turner.
\newblock Meta-learning stationary stochastic process prediction with
  convolutional neural processes.
\newblock \emph{Advances in Neural Information Processing Systems},
  33:\penalty0 8284--8295, 2020.

\bibitem[Garnelo et~al.(2018{\natexlab{a}})Garnelo, Rosenbaum, Maddison,
  Ramalho, Saxton, Shanahan, Teh, Rezende, and Eslami]{garnelo2018conditional}
Marta Garnelo, Dan Rosenbaum, Christopher Maddison, Tiago Ramalho, David
  Saxton, Murray Shanahan, Yee~Whye Teh, Danilo Rezende, and SM~Ali Eslami.
\newblock Conditional neural processes.
\newblock In \emph{International Conference on Machine Learning}, pp.\
  1704--1713. PMLR, 2018{\natexlab{a}}.

\bibitem[Garnelo et~al.(2018{\natexlab{b}})Garnelo, Schwarz, Rosenbaum, Viola,
  Rezende, Eslami, and Teh]{garnelo2018neural}
Marta Garnelo, Jonathan Schwarz, Dan Rosenbaum, Fabio Viola, Danilo~J Rezende,
  SM~Eslami, and Yee~Whye Teh.
\newblock Neural processes.
\newblock \emph{arXiv preprint arXiv:1807.01622}, 2018{\natexlab{b}}.

\bibitem[Gordon et~al.()Gordon, Bruinsma, Foong, Requeima, Dubois, and
  Turner]{Gordon2020convolutional}
Jonathan Gordon, Wessel~P. Bruinsma, Andrew Y.~K. Foong, James Requeima, Yann
  Dubois, and Richard~E. Turner.
\newblock Convolutional conditional neural processes.
\newblock In \emph{8th International Conference on Learning Representations,
  {ICLR} 2020}.

\bibitem[Higgins et~al.()Higgins, Matthey, Pal, Burgess, Glorot, Botvinick,
  Mohamed, and Lerchner]{irina2017beta}
Irina Higgins, Lo{\"{\i}}c Matthey, Arka Pal, Christopher~P. Burgess, Xavier
  Glorot, Matthew~M. Botvinick, Shakir Mohamed, and Alexander Lerchner.
\newblock beta-vae: Learning basic visual concepts with a constrained
  variational framework.
\newblock In \emph{5th International Conference on Learning Representations,
  {ICLR} 2017}.

\bibitem[Ivanov et~al.(2019)Ivanov, Figurnov, and
  Vetrov]{ivanov2019variational}
Oleg Ivanov, Michael Figurnov, and Dmitry~P. Vetrov.
\newblock Variational autoencoder with arbitrary conditioning.
\newblock In \emph{7th International Conference on Learning Representations,
  {ICLR} 2019}, 2019.

\bibitem[Jha et~al.(2022)Jha, Gong, Wang, Turner, and Yao]{jha2022neural}
Saurav Jha, Dong Gong, Xuesong Wang, Richard~E Turner, and Lina Yao.
\newblock The neural process family: Survey, applications and perspectives.
\newblock \emph{arXiv preprint arXiv:2209.00517}, 2022.

\bibitem[Karras et~al.(2020)Karras, Laine, Aittala, Hellsten, Lehtinen, and
  Aila]{karras2020analyzing}
Tero Karras, Samuli Laine, Miika Aittala, Janne Hellsten, Jaakko Lehtinen, and
  Timo Aila.
\newblock Analyzing and improving the image quality of stylegan.
\newblock In \emph{Proceedings of the IEEE/CVF conference on computer vision
  and pattern recognition}, pp.\  8110--8119, 2020.

\bibitem[Karras et~al.(2021)Karras, Aittala, Laine, H{\"a}rk{\"o}nen, Hellsten,
  Lehtinen, and Aila]{karras2021alias}
Tero Karras, Miika Aittala, Samuli Laine, Erik H{\"a}rk{\"o}nen, Janne
  Hellsten, Jaakko Lehtinen, and Timo Aila.
\newblock Alias-free generative adversarial networks.
\newblock volume~34, 2021.

\bibitem[Kim et~al.(2019)Kim, Mnih, Schwarz, Garnelo, Eslami, Rosenbaum,
  Vinyals, and Teh]{kim2019attentive}
Hyunjik Kim, Andriy Mnih, Jonathan Schwarz, Marta Garnelo, S.~M.~Ali Eslami,
  Dan Rosenbaum, Oriol Vinyals, and Yee~Whye Teh.
\newblock Attentive neural processes.
\newblock In \emph{7th International Conference on Learning Representations,
  {ICLR} 2019}, 2019.

\bibitem[Kim et~al.(2022)Kim, Go, and Yun]{kim2022neural}
Mingyu Kim, Kyeong~Ryeol Go, and Se-Young Yun.
\newblock Neural processes with stochastic attention: Paying more attention to
  the context dataset.
\newblock In \emph{10th International Conference on Learning Representations,
  {ICLR} 2022}, 2022.

\bibitem[Kingma \& Welling(2014)Kingma and Welling]{kingma2014auto}
Diederik~P. Kingma and Max Welling.
\newblock Auto-encoding variational bayes.
\newblock In \emph{2nd International Conference on Learning Representations,
  {ICLR} 2014}, 2014.

\bibitem[Kosiorek et~al.(2021)Kosiorek, Strathmann, Zoran, Moreno, Schneider,
  Mokr{\'a}, and Rezende]{kosiorek2021nerf}
Adam~R Kosiorek, Heiko Strathmann, Daniel Zoran, Pol Moreno, Rosalia Schneider,
  Sona Mokr{\'a}, and Danilo~Jimenez Rezende.
\newblock Nerf-vae: A geometry aware 3d scene generative model.
\newblock In \emph{International Conference on Machine Learning}, pp.\
  5742--5752. PMLR, 2021.

\bibitem[Lee et~al.(2021)Lee, Tack, Lee, and Shin]{lee2021meta}
Jaeho Lee, Jihoon Tack, Namhoon Lee, and Jinwoo Shin.
\newblock Meta-learning sparse implicit neural representations.
\newblock volume~34, pp.\  11769--11780, 2021.

\bibitem[Lee et~al.(2020)Lee, Lee, Kim, Yang, Hwang, and
  Teh]{lee2020bootstrapping}
Juho Lee, Yoonho Lee, Jungtaek Kim, Eunho Yang, Sung~Ju Hwang, and Yee~Whye
  Teh.
\newblock Bootstrapping neural processes.
\newblock volume~33, pp.\  6606--6615, 2020.

\bibitem[Liu et~al.(2015)Liu, Luo, Wang, and Tang]{liu2015deep}
Ziwei Liu, Ping Luo, Xiaogang Wang, and Xiaoou Tang.
\newblock Deep learning face attributes in the wild.
\newblock In \emph{Proceedings of the IEEE international conference on computer
  vision}, pp.\  3730--3738, 2015.

\bibitem[Martin-Brualla et~al.(2021)Martin-Brualla, Radwan, Sajjadi, Barron,
  Dosovitskiy, and Duckworth]{martin2021nerf}
Ricardo Martin-Brualla, Noha Radwan, Mehdi~SM Sajjadi, Jonathan~T Barron,
  Alexey Dosovitskiy, and Daniel Duckworth.
\newblock Nerf in the wild: Neural radiance fields for unconstrained photo
  collections.
\newblock In \emph{Proceedings of the IEEE/CVF Conference on Computer Vision
  and Pattern Recognition}, pp.\  7210--7219, 2021.

\bibitem[Mildenhall et~al.(2020)Mildenhall, Srinivasan, Tancik, Barron,
  Ramamoorthi, and Ng]{mildenhall2020nerf}
Ben Mildenhall, Pratul~P Srinivasan, Matthew Tancik, Jonathan~T Barron, Ravi
  Ramamoorthi, and Ren Ng.
\newblock Nerf: Representing scenes as neural radiance fields for view
  synthesis.
\newblock In \emph{European conference on computer vision}, pp.\  405--421.
  Springer, 2020.

\bibitem[Nguyen \& Grover(2022)Nguyen and Grover]{nguyen2022transformer}
Tung Nguyen and Aditya Grover.
\newblock Transformer neural processes: Uncertainty-aware meta learning via
  sequence modeling.
\newblock In \emph{International Conference on Machine Learning}, pp.\
  16569--16594. PMLR, 2022.

\bibitem[Niemeyer \& Geiger(2021)Niemeyer and Geiger]{niemeyer2021giraffe}
Michael Niemeyer and Andreas Geiger.
\newblock Giraffe: Representing scenes as compositional generative neural
  feature fields.
\newblock In \emph{Proceedings of the IEEE/CVF Conference on Computer Vision
  and Pattern Recognition}, pp.\  11453--11464, 2021.

\bibitem[Oechsle et~al.(2019)Oechsle, Mescheder, Niemeyer, Strauss, and
  Geiger]{oechsle2019texture}
Michael Oechsle, Lars Mescheder, Michael Niemeyer, Thilo Strauss, and Andreas
  Geiger.
\newblock Texture fields: Learning texture representations in function space.
\newblock In \emph{Proceedings of the IEEE/CVF International Conference on
  Computer Vision}, pp.\  4531--4540, 2019.

\bibitem[Park et~al.(2019)Park, Florence, Straub, Newcombe, and
  Lovegrove]{park2019deepsdf}
Jeong~Joon Park, Peter Florence, Julian Straub, Richard Newcombe, and Steven
  Lovegrove.
\newblock Deepsdf: Learning continuous signed distance functions for shape
  representation.
\newblock In \emph{Proceedings of the IEEE/CVF Conference on Computer Vision
  and Pattern Recognition}, pp.\  165--174, 2019.

\bibitem[Ross et~al.(1996)Ross, Kelly, Sullivan, Perry, Mercer, Davis,
  Washburn, Sager, Boyce, and Bristow]{ross1996stochastic}
Sheldon~M Ross, John~J Kelly, Roger~J Sullivan, William~James Perry, Donald
  Mercer, Ruth~M Davis, Thomas~Dell Washburn, Earl~V Sager, Joseph~B Boyce, and
  Vincent~L Bristow.
\newblock \emph{Stochastic processes}, volume~2.
\newblock Wiley New York, 1996.

\bibitem[Salimans et~al.()Salimans, Karpathy, Chen, and
  Kingma]{pixelcnn2017tim}
Tim Salimans, Andrej Karpathy, Xi~Chen, and Diederik~P. Kingma.
\newblock Pixelcnn++: Improving the pixelcnn with discretized logistic mixture
  likelihood and other modifications.
\newblock In \emph{5th International Conference on Learning Representations,
  {ICLR} 2017,}.

\bibitem[Schwarz \& Teh(2022)Schwarz and Teh]{schwarz2022metalearning}
Jonathan Schwarz and Yee~Whye Teh.
\newblock Meta-learning sparse compression networks.
\newblock \emph{Transactions on Machine Learning Research}, 2022.
\newblock ISSN 2835-8856.

\bibitem[Sitzmann et~al.(2020{\natexlab{a}})Sitzmann, Chan, Tucker, Snavely,
  and Wetzstein]{sitzmann2020metasdf}
Vincent Sitzmann, Eric Chan, Richard Tucker, Noah Snavely, and Gordon
  Wetzstein.
\newblock Metasdf: Meta-learning signed distance functions.
\newblock volume~33, pp.\  10136--10147, 2020{\natexlab{a}}.

\bibitem[Sitzmann et~al.(2020{\natexlab{b}})Sitzmann, Martel, Bergman, Lindell,
  and Wetzstein]{sitzmann2020implicit}
Vincent Sitzmann, Julien Martel, Alexander Bergman, David Lindell, and Gordon
  Wetzstein.
\newblock Implicit neural representations with periodic activation functions.
\newblock volume~33, pp.\  7462--7473, 2020{\natexlab{b}}.

\bibitem[Sohn et~al.(2015)Sohn, Lee, and Yan]{sohn2015learning}
Kihyuk Sohn, Honglak Lee, and Xinchen Yan.
\newblock Learning structured output representation using deep conditional
  generative models.
\newblock volume~28, 2015.

\bibitem[S{\o}nderby et~al.(2016)S{\o}nderby, Raiko, Maal{\o}e, S{\o}nderby,
  and Winther]{sonderby2016ladder}
Casper~Kaae S{\o}nderby, Tapani Raiko, Lars Maal{\o}e, S{\o}ren~Kaae
  S{\o}nderby, and Ole Winther.
\newblock Ladder variational autoencoders.
\newblock \emph{Advances in neural information processing systems}, 29, 2016.

\bibitem[Stanley(2007)]{stanley2007compositional}
Kenneth~O Stanley.
\newblock Compositional pattern producing networks: A novel abstraction of
  development.
\newblock \emph{Genetic programming and evolvable machines}, 8\penalty0
  (2):\penalty0 131--162, 2007.

\bibitem[Tancik et~al.(2020)Tancik, Srinivasan, Mildenhall, Fridovich-Keil,
  Raghavan, Singhal, Ramamoorthi, Barron, and Ng]{tancik2020fourier}
Matthew Tancik, Pratul Srinivasan, Ben Mildenhall, Sara Fridovich-Keil, Nithin
  Raghavan, Utkarsh Singhal, Ravi Ramamoorthi, Jonathan Barron, and Ren Ng.
\newblock Fourier features let networks learn high frequency functions in low
  dimensional domains.
\newblock volume~33, pp.\  7537--7547, 2020.

\bibitem[Vahdat \& Kautz(2020)Vahdat and Kautz]{vahdat2020nvae}
Arash Vahdat and Jan Kautz.
\newblock Nvae: A deep hierarchical variational autoencoder.
\newblock volume~33, pp.\  19667--19679, 2020.

\bibitem[Vaswani et~al.(2017)Vaswani, Shazeer, Parmar, Uszkoreit, Jones, Gomez,
  Kaiser, and Polosukhin]{vaswani2017attention}
Ashish Vaswani, Noam Shazeer, Niki Parmar, Jakob Uszkoreit, Llion Jones,
  Aidan~N Gomez, {\L}ukasz Kaiser, and Illia Polosukhin.
\newblock Attention is all you need.
\newblock volume~30, 2017.

\bibitem[Volpp et~al.(2021)Volpp, Fl{\"u}renbrock, Grossberger, Daniel, and
  Neumann]{volpp2021bayesian}
Michael Volpp, Fabian Fl{\"u}renbrock, Lukas Grossberger, Christian Daniel, and
  Gerhard Neumann.
\newblock Bayesian context aggregation for neural processes.
\newblock In \emph{9th International Conference on Learning Representations,
  {ICLR} 2021}, 2021.

\bibitem[Wang \& Van~Hoof(2020)Wang and Van~Hoof]{wang2020doubly}
Qi~Wang and Herke Van~Hoof.
\newblock Doubly stochastic variational inference for neural processes with
  hierarchical latent variables.
\newblock In \emph{International Conference on Machine Learning}, pp.\
  10018--10028. PMLR, 2020.

\bibitem[Zaheer et~al.(2017)Zaheer, Kottur, Ravanbakhsh, Poczos, Salakhutdinov,
  and Smola]{zaheer2017deep}
Manzil Zaheer, Satwik Kottur, Siamak Ravanbakhsh, Barnabas Poczos, Russ~R
  Salakhutdinov, and Alexander~J Smola.
\newblock Deep sets.
\newblock volume~30, 2017.

\end{thebibliography}
\bibliographystyle{iclr2023_conference}

\newpage

\appendix

\section*{Appendix A: More Details on Modulated MLP Layer}

The modulated MLP layer used in our paper is similar to the style modulation in \cite{karras2020analyzing}. Mathematically, we denote the weights of a MLP layer (1x1 convolution) by $W \in \mathbb{R}^{d_{in} \times d_{out}}$, where $d_{in}$, $d_{out}$, and $ w_{ij}$ denote the dimension of input and output, the element in $i^{th}$ row and $j^{th}$ column of $W$, respectively. We obtain the style vector $\mathbf{s} \in \mathbb{R}^{d_{in}}$ by passing the latent variable $\mathbf{z}$ into two MLP layers. The $i^{th}$ element of style vector $s_i$ is thus used to modulate the parameter of $W$ as, 
\begin{equation}
    w'_{ij} = s_i \cdot w_{ij}, \hspace{0.4cm}j=1,\cdots, d_{out},
\end{equation}
where $w_{ij}$ and $w'_{ij}$ denote the original and modulated weights, respectively.  

%The style vector (in the shape of $1\times d$) is obtained as the result after passing every latent variable into MLPs. Note the process is implemented with group convolution. 
The modulated weights are normalized to preserve training stability,
\begin{equation}
    w''_{ij} = w'_{ij} \  / \sqrt{\sum\limits_{i}{w_{ij}^{'2}+\epsilon}}, \hspace{0.4cm}j=1,\cdots, d_{out},
\end{equation}
where $\epsilon$ is a very small constant to prevent the denominator to be zero. These two equations describe the mechanism of the modulate MLP used in our method.
%It can adaptively adjust the parameter weights $w'$ in MLPs according to the sampled latent variable.
% The modulated MLP layer used in our paper is similar to the style modulation in \cite{karras2020analyzing}. Specifically, the weights of MLP (1x1 convolution, input dim is $d$ and output dim is $d$) are scaled by the style vector, 
% \begin{equation}
%     w'_{ij} = s_i \cdot w_{ij}, 
% \end{equation}
% where $w$ and $w'$ denote the original and modulated weights, respectively. $s_i$ is the scale corresponding to the $i^{th}$ input style vector ($i=1,2,...d$, $d$ is the channel number). $j$ enumerates the output feature maps of the convolution. 

% The style vector (in the shape of $1\times d$) is obtained as the result after passing every latent variable into MLPs. Note the process is implemented with group convolution. After modulation and convolution, the output is normalized to preserve training stability,
% \begin{equation}
%     w''_{ij} = w'_{ij} \  / \sqrt{\sum\limits_{i}{w'_{ij}}+\epsilon},
% \end{equation}
% where $\epsilon$ is a very small constant to prevent the denominator to be zero. These two equations describe the mechanism of the modulate MLP used in our method. It can adaptively adjust the parameter weights $w'$ in MLPs according to the sampled latent variable.

\section*{Appendix B: Detailed Experimental Settings}
\label{appendix_a}

Here, we provide the detailed hyper parameters used in all the experiments in Table \ref{table4}. Among them, some hyper parameters denote to, $L_s$ number of self-attention layers in bottleneck encoder, $L_c$ number of cross-attention layers, $K$ number of hierarchical blocks, $d$ the feature dimension through the entire network. The row of Fourier features describes whether we embed the coordinate into high-frequency embeddings as suggested in \citep{tancik2020fourier}. $\beta$ is the hyper parameter used to balance the losses of reconstruction and KL divergence. The reconstruction term describes how we calculate the reconstruction loss  $\mathbb{E}_{z \sim q_{\phi}(z|D_T)} [- \log p_{\theta}(Y_T |  z, X_T, D_C)].$

\newcommand{\cmark}{\ding{51}}%
\newcommand{\xmark}{\ding{55}}%

\begin{table}[h]
\renewcommand{\arraystretch}{1.2}
\begin{tabular}{cccc}
\hline
                    & 1D synthetic functions & 2D images                & 3D scenes    \\ \hline
$L_s$        & 3                      & 6                        & 6            \\ \hline
$L_c$                & 1                      & 4                        & 2            \\ \hline
$L_K$                   & 6                      & 6                        & 4            \\ \hline
$d$                  & 128                    & 512                      & 512          \\ \hline
$d_z$                  & 16                     & 64                       & 64           \\ \hline
model size          & 2.3M                   & 56.7M                    & 34.3M        \\ \hline
Fourier features    & \xmark                 & \cmark                   & \cmark      \\ \hline
batch size          & 100                    & 16                       & 32           \\ \hline
iteration number          & 0.1 million            & 0.5 million              & 0.1 million  \\ \hline
lr                  & 5e-4                   & 1e-4                     & 1e-4         \\ \hline
$\beta$                & 1                      & 0.1                     & 0.001        \\ \hline
\multirow{2}{*}{reconstruction term} & \multirow{2}{*}{Gaussian} & discrete logistic mixture               & \multirow{2}{*}{MSE} \\
                                     &                           & \citep{pixelcnn2017tim}                       &                      \\ \hline
Training resources        & 1x V100 16GB           & 4x V100 16GB (CelebA178) & 4x V100 32GB \\ \hline
\end{tabular}
\caption{Hyper parameters in our experiments. \label{table4}}
\end{table}

In addition, when training for 1D synthetic functions, we will randomly sample 5 to 15 context points and 15 to 25 target points on Matern and RBF kernel. For the last four rows listed in the table 1 of our main paper, we use importance weighted sampling to calculate the approximated log likelihood. We use this metric because these four models are all latent variable models, where the results of importance weighted sampling would be more accurate, especially for hierarchical latent variable models such as NVAE \citep{vahdat2020nvae}.
 
Training a VNP model for the 1D regression task requires about 5 hours. Training a 2D VNP model on CelebA64 requires about 24 hours. Training a 3D VNP model for novel view synthesis takes around 40 hours. For the inference speed, VNP requires very short time for 1D and 2D tasks. On 1D regression task, VNP takes 0.285 second for testing a batch with batch size of 2000. On the CelebA64 dataset (2D task), our VNP takes around 0.112 second to infer a single batch of size 8. On the 3D Cars dataset, our VNP takes 5.28 second to render an image with a novel view (reasons explained in Appendix C). All these results are tested with a single V100 GPU. 
% Training a VNP model for the 1D regression task requires about 5 hours. Training a 2D VNP model on CelebA64 requires about 24 hours. Training a 3D VNP model for novel view synthesis requires about 40 hours. For the inference speed, VNP requires very short time for 1D and 2D tasks. On 1D regression task, the proposed VNP requires 0.285 for test a batch with batch size as 2000. On CelebA64 dataset, our VNP requires around 0.112 s to infer with a single batch (batch size=8). On 3D Cars dataset, our VNP takes 5.28 s to render an image with a novel view as explained in Appendix C. All these results are tested with a single V100 GPU.

\section*{Appendix C: Comparisons with the Optimization-based Method}

In this section, we provide the comparison of our VNP and the previous optimization-based method. We take a representative optimization-based method, SIREN \citep{sitzmann2020implicit} for comparison. We conduct experiments on the 3D Car dataset \cite{chang2015shapenet}, and measure both the inference speed and the PSNR of the novel synthesized view. Note that since SIREN requires to optimize the network to fit for a specific signal, the iteration time of SIREN is a part of the inference time. We test the inference speed with a single V100 GPU for both schemes.

The results are shown in Table \ref{appendix_siren}. It is observed that for the task of novel view synthesis conditioned on a single view, our proposed VNP provides much better prediction performance compared with the optimized-based method SIREN, even if SIREN is optimized for many iterations for a given test image. One reason is that VNP can learn the dataset prior, which complements many useful information to predict a specific 3D signal. However, if we optimize the SIREN network to fit the known single view, the network will be initially optimized for the right direction, but will then tend to overfit to this single view after many iterations. As a result, it is observed that when the SIREN network gets better prediction performance as the iteration number increased from 1 to 100. However, the performance decreases if we apply more iterations (such as 300 iterations), compared with that of 100 iterations.

As for the inference speed, we can see that our VNP takes 5.28s to render an image in the novel view with a single forward pass. The reason for such a long inference time is that we use a large number of sampling points in 3D space. If an image is with the resolution of $128 \times 128$, to render the RGB value of every pixel, there are $128 \times 128 \times p$ target points, where $p$ is the number of sampled point along each ray (set as $p=128$ in our experiment). Therefore, processing all these target points with cross-attention requires huge GPU memory. We have to divide the image into several pixel groups, and render these pixel groups sequentially. Currently, we have not implement any optimization of our code. We will improve the implementation of method on tasks involving 3D signals in the future.

Note our VNP only presents slow inference speed on this 3D novel view synthesis task. 
On the 1D and 2D tasks, since there are not so many target points and the model does not incorporate rendering process, the inference speed of our method is fast.

\begin{table}[h]
\centering
\renewcommand{\arraystretch}{1.2}
\begin{tabular}{|cccccc|c|}
\hline
\multicolumn{6}{|c|}{SIREN}                                                                                                           & VNP (4 blocks)          \\ \hline
\multicolumn{1}{|c|}{Iteration Number} & \multicolumn{1}{c|}{1}     & \multicolumn{1}{c|}{10}    & \multicolumn{1}{c|}{30} & \multicolumn{1}{c|}{100}   & 300  & no finetuning \\ \hline
\multicolumn{1}{|c|}{Time (s)}         & \multicolumn{1}{c|}{0.21}  & \multicolumn{1}{c|}{1.35} & \multicolumn{1}{c|}{3.68} & \multicolumn{1}{c|}{11.59}  & 33.68   &  4.79       \\ \hline
\multicolumn{1}{|c|}{PSNR}             &  \multicolumn{1}{c|}{11.92} & \multicolumn{1}{c|}{12.27} &  \multicolumn{1}{c|}{12.72} & \multicolumn{1}{c|}{12.73} & 12.00   &  24.21         \\ \hline
\end{tabular}
\caption{Novel view synthesis conditioned on a single view. We evaluate the inference time and the prediction performance (PSNR) on ShapeNet Cars \citep{chang2015shapenet}. \label{appendix_siren}}
\end{table}

\section*{Appendix D: Limitation}

The proposed VNP inherits the limitation of NP family. NPs are interesting techniques for meta-learning implicit neural representations due to their reduction of the high cost of training. NPs learns the common knowledge shared by the dataset, enabling fast inference of an unseen signal without the need of finetuning. NPs still cannot work well for dataset including diverse objects (with less shared knowledge), e.g., ImageNet.

\end{document}